\documentclass[10pt,twocolumn,letterpaper]{article}

\usepackage[pagenumbers]{iccv} % To force page numbers, e.g. for an arXiv version

\newcommand{\head}[1]{{\noindent\textbf{#1}}}

\newcommand{\model}{{\scshape Sail}\xspace}
\newcommand{\boldmodel}{{\scshape \textbf{Sail}}\xspace}

\usepackage{tcolorbox}
\tcbuselibrary{most}
\tcbset{
  aibox/.style={
    width=\textwidth,
    top=10pt,
    colback=white,
    colframe=black,
    colbacktitle=black,
    enhanced,
    center,
    attach boxed title to top left={yshift=-0.1in,xshift=0.15in},
    boxed title style={boxrule=0pt,colframe=white,},
  }
}
\newtcolorbox{AIbox}[2][]{aibox,title=#2,#1}

\usepackage{fancyhdr}
\definecolor{iccvblue}{rgb}{0.21,0.49,0.74}
\usepackage[pagebackref,breaklinks,colorlinks,allcolors=iccvblue]{hyperref}
\usepackage{multicol}
\usepackage{multirow}
\usepackage{booktabs}
\usepackage{makecell}
\usepackage{array}
\title{
    The Scalability of Simplicity:\\
    Empirical Analysis of Vision-Language Learning with a Single Transformer
}

\author {
    Weixian Lei\textsuperscript{\rm *}\quad
    Jiacong Wang\textsuperscript{\rm *}\quad
    Haochen Wang\textsuperscript{\rm *}\quad
    \\
    Xiangtai Li\quad
    Jun Hao Liew\quad
    Jiashi Feng\quad
    Zilong Huang$^{\dag}$\\
    \small $^{*}$Equal contribution, $^\dag$ Project Lead \\
    Bytedance Seed
}

\begin{document}

\maketitle

\thispagestyle{fancy}
\fancyhead[L]{\includegraphics[height=15pt]{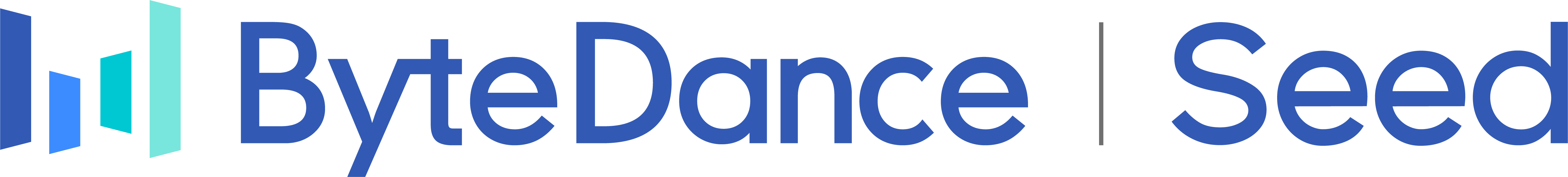}}
\fancyhead[R]{}

\begin{abstract}
This paper introduces \model,  a single transformer  unified multimodal large language model (MLLM) that integrates   raw pixel encoding and language decoding within a singular architecture. 
Unlike existing modular MLLMs, which rely on a pre-trained vision transformer (ViT), \model eliminates the need for a separate vision encoder, presenting a more minimalist architecture design.
Instead of introducing novel architectural components, \model adapts mix-attention mechanisms and multimodal positional encodings to better align with the distinct characteristics of visual and textual modalities. 
We systematically  compare \model's properties\textemdash including scalability, cross-modal information flow patterns, and visual representation capabilities\textemdash  with those of modular MLLMs. 
By scaling both training data and model size, \model achieves performance comparable to modular MLLMs. 
Notably, the removal of pretrained ViT 
components enhances \model's  scalability and results in significantly  different cross-modal information flow patterns. 
Moreover, \model demonstrates strong visual representation capabilities, achieving results on par with ViT-22B in vision tasks such as semantic segmentation.
Code and models are available\footnote{\url{https://github.com/bytedance/SAIL}}.
\end{abstract}

% replace SAIL with \model    
\section{Introduction}
\label{sec:intro}

The pursuit of multimodal intelligence has driven the development of Multimodal Large Language Models (MLLMs)~\cite{gpt4v,gemini,llava}, which typically adopt a modular design: a pre-trained vision encoder (\eg, CLIP-ViT~\cite{openai_clip,openclip}) extracts image features, a Large Language Model (LLM)~\cite{chatgpt,gpt4,claude,llama3,vicuna2023,jiang2023mistral} processes text, and a lightweight projector aligns the two modalities. 
This framework achieves strong performance through multi-stage pretraining, supervised fine-tuning (SFT), and post-training on multimodal datasets~\cite{llava,zhu2023minigpt,dai2023instructblip,wang2024world,bai2023qwen-vl,qwen2vl}. 
While effective, this modular MLLM paradigm inherently fragments multimodal processing, reinforces reliance on pretrained visual encoders, which  may limit  deployment flexibility and scalability~\cite{diao2024EVE,chen2024SOLO,luo2024mono-internvl}.

\begin{figure}[t]
    \centering
    \includegraphics[width=1.0\columnwidth]{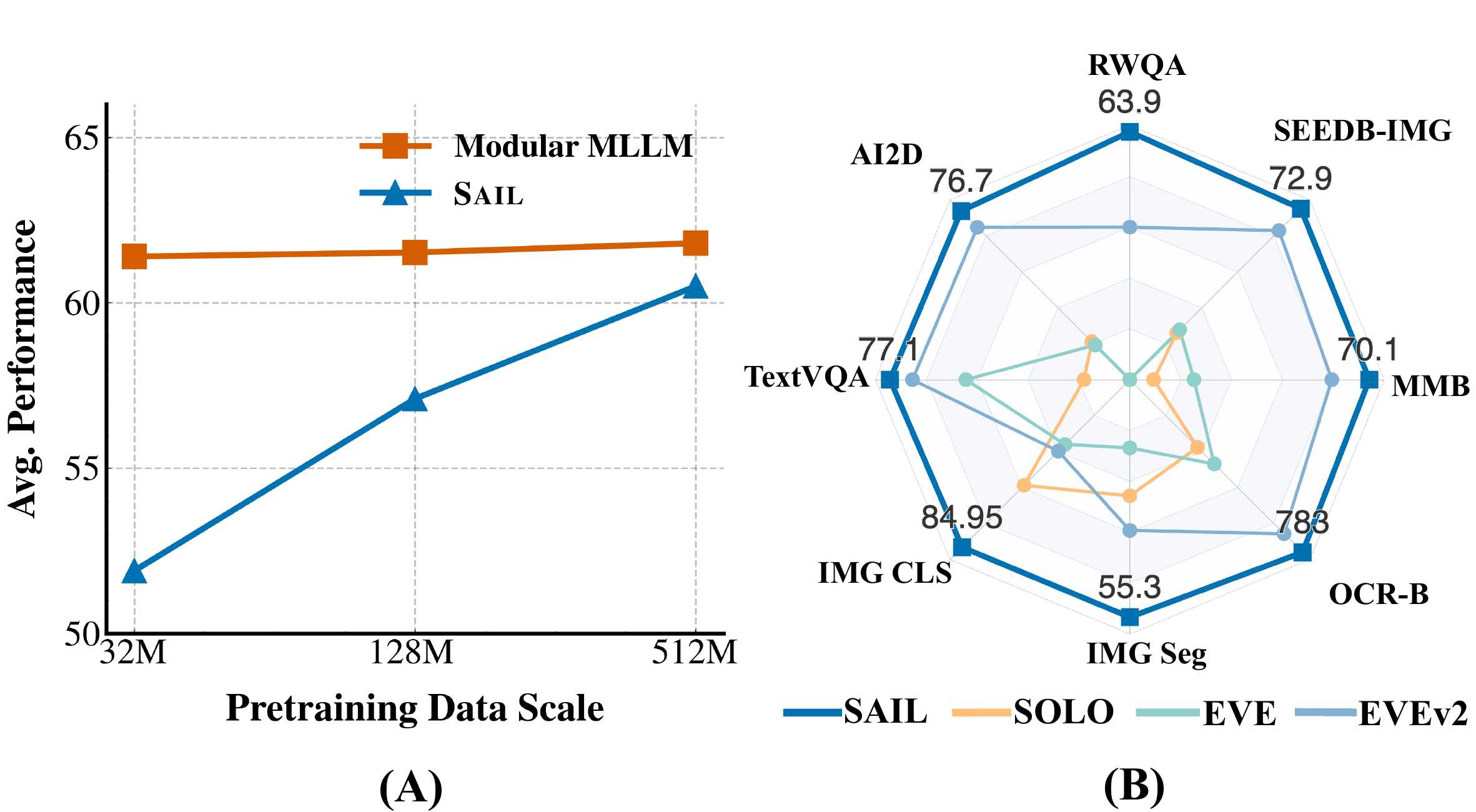}
    % \vspace{-0.5em}
\caption{
    \textbf{(A) Data scaling curve} for Modular Multimodal Large Language Model (MLLM) and \model, our Single Transformer-based MLLM. As pretraining data increases, the single transformer \model shows a sharper performance gain, demonstrating  its superior data scalability.
    \textbf{(B) Comparison to existing Single Transformer-based MLLMs}: our \model pushes the performance boundaries on both vision tasks and vision-language tasks.
    }
    \label{fig:teaser}
    \vspace{-1.5em}
\end{figure}

A promising alternative is to eliminate the visual encoder entirely and process raw image patches and text tokens within a single Transformer. 
This unified architecture removes modality-specific modules, enabling parameter sharing and end-to-end learning of vision-language interactions. 
Previous works~\cite{diao2024EVE,chen2024SOLO,luo2024mono-internvl} have  primarily   explored the architecture design, training data, and methods of Single Transformer-based MLLMs. 
However,  little exploration has been given to their fundamental properties, such as   \textit{scalability}, \textit{cross-modal information flow patterns}, and \textit{visual representation capabilities}. 
A deeper understanding of these properties is crucial  for unlocking the full potential of Single Transformer-based MLLMs.

In this work, we present an experimental analysis of  the   fundamental  properties of Single Transformer-based MLLMs and compare them to    modular MLLMs (\eg LLaVA~\cite{llava}).  
Addtionally, in the absence of a pre-trained visual encoder, Single Transformers have to learn visual representations from scratch. 
Thus, an intriguing question arises: can  a trained Single Transformer   emerge as a strong vision encoder?

We conduct a series of experiments to train and study our \textbf{S}ingle tr\textbf{A}nsformer model for v\textbf{I}sion and \textbf{L}anguage (\boldmodel). While we do not propose novel architecture designs, we introduce necessary modifications to enable the model to process different modalities in a unified architecture.  In its micro architecture design, we address the different spatial characteristics of 2D images and 1D text data by employing a mixed attention mechanism: bidirectional attention for image patches and causal attention for text tokens, combined with multimodal rotary position embedding. 
Through model   and data scaling,   \model   achieves performance on vision-language benchmarks comparable to modular MLLMs, while also functioning as a high-performing vision backbone, as shown in Figure~\ref{fig:teaser}.

More concretely, our empirical analysis uncovers following striking advantages of Single Transformer architectures:

\head{\textit{(i)} Superior Data Scaling:} In controlled experiments, \model   exhibits steeper performance gains as pretraining data scales. While LLaVA-style modular MLLMs initially perform well, our model's performance becomes very close to theirs when pretrained on 512M samples, as shown in Figure~\ref{fig:teaser}(A). This suggests that unified architectures can effectively leverage large-scale data and potentially match the performance of modular MLLMs.

\head{\textit{(ii)} Vision Centric Information Flow Pattern:} 
Through analysis of attention distributions, we observe that Single Transformers assign significantly higher attention scores to image tokens during token prediction compared to modular MLLMs. This indicates that the information flow in Single Transformer MLLMs is more direct, with visual tokens influencing prediction tokens more prominently, highlighting  a vision-centric approach to decision-making.

\head{\textit{(iii)} Vision Encoder Functioning:} Our experiments further demonstrate  that the pretrained Single Transformer inherently serves as a powerful vision encoder. Comprehensive evaluations on vision-centric tasks, such as image classification and semantic segmentation, show that the model learns rich visual representations during multimodal pretraining. These representations enhance its capacity for both semantic-level comprehension (\eg, object categorization) and pixel-level understanding (\eg, fine-grained segmentation masks), bridging high-level abstraction and low-level visual reasoning within a unified architecture.

In summary, our findings indicate that Single Transformer-based MLLMs hold great promise in surpassing modular MLLMs in terms of leveraging large-scale data, forming direct vision-centric information pathways, and functioning as effective vision encoders. 
We hope our empirical findings inspire further research to refine and enhance Single Transformer architecture, ultimately driving advancements in multimodal intelligence from a new perspective.

\section{Related Work}
\label{sec:related_work}
\subsection{Paradigms in Vision-Language Model Design}
\head{Modular MLLMs with Visual Encoders.}
The prevailing approach in MLLM design employs modular architectures~\cite{gpt4v,gemini,llava,bai2023qwen-vl,qwen2vl} that rely on  pretrained vision encoders (\eg, CLIP-ViT~\cite{openai_clip,openclip}, InternViT~\cite{chen2024internvl}) to process visual inputs.  
The visual features extracted from these frozen encoders are then aligned with LLM input spaces via linear~\cite{llava,zhu2023minigpt,chen2023minigptv2, wang2025ross} or cross-attention layers~\cite{alayrac2022flamingo,cogagent}. 
While this module design enables effective transfer of   pretrained visual-language knowledge, it also introduces several limitations. 
First, incorporating a  separate ViT encoder significantly slows down both training and inference, increasing deployment complexity and requiring   costly infrastructure\textemdash especially when compared to a single transformer unified model. Second, common strategies for integrating visual features, such as direct   mapping into LLM inputs~\cite{llava,zhu2023minigpt,chen2023minigptv2} or sharing them across LLM layers~\cite{cogagent,alayrac2022flamingo}, often struggle to reconcile the inherent differences between images and text representations. Finally, as model scale,  balancing the interactions between the  the encoder, LLM, and alignment layers becomes increasingly challenging~\cite{chen2024SOLO,luo2024mono-internvl}. Thus, in this work, we  explore a single transformer-based MLLM architecture that eliminates the ViT encoder and alignment components to overcome these challenges.  %\lxt{We can shorten this part, since this is a little repeated with the first paragraph in the introduction part.}
%\lxt{We should add one sentence to introduce our goal at the end of paragraph.}

\head{Single Transformer-based MLLMs Without Visual Encoders.}
Emerging research explores end-to-end architectures that process raw image patches and text tokens through a single Transformer, bypassing visual encoders entirely. These monolithic designs fall into two categories: continuous tokenization and discrete tokenization. Continuous tokenization, exemplified by Fuyu-8B~\cite{fuyu-8b} and SOLO~\cite{chen2024SOLO}, directly maps patches to LLM embeddings via linear projections, enabling flexible resolution handling but requiring massive pretraining data. Discrete tokenization, adopted by Chameleon~\cite{chameleon} and Emu3~\cite{emu3}, employs VQ-VAE tokenizers to compress images into discrete tokens, trading pixel-level fidelity for generation capabilities. 
While later efforts such as EVE~\cite{diao2024EVE} and Mono-InternVL~\cite{luo2024mono-internvl} demonstrate the feasibility of encoder-free training, critical gaps remain: 
(1) Existing methods rely on   extra designs and auxiliary loss~\cite{diao2024EVE}, complicating training pipelines; 
(2) The scaling laws and fundamental properties of purely end-to-end trained models remain poorly understood; 
(3) Vision-language interaction   in shared parameter spaces lacks systematic analysis\textemdash most prior MLLMs default to   causal attention for processing image-text sequences. 
In this work, we   reveal that enabling bidirectional attention between image patches significantly enhances visual representation learning, addressing a key limitation   in current designs.
More importantly, our study bridges these gaps by establishing foundational principles for training scalable, self-contained single-transformer MLLMs.

\subsection{Vision Representation Learning}
Learning  effective vision representations is a core challenge in computer vision research, with extensive works~\cite{deng2009imagenet,dehghani2023scaling,wang2022simvlm,fang2023eva,bao2021beit,he2020momentum} dedicated to  this problem. With the proliferation of large-scale web-sourced image-text datasets~\cite{cc12m,datacomp1b,schuhmann2022laion5b}, recent methods leverage this data to train deep visual representations via three primary paradigms: 

\head{Text as Classification Labels.}
Early methods used textual descriptions as weak supervision by extracting categorical labels from captions. For example, frameworks like Tag2Text~\cite{huang2023tag2text} and RAM~\cite{RAM} used ViTs \cite{dosovitskiy2021vit} to predict noun-based pseudo-labels from datasets like CC12M~\cite{cc12m}. CatLIP~\cite{mehta2024catlip} scaled labels to millions using object-centric supervision, and SuperClass~\cite{superclass_huang} directly used tokenized text tokens as classification categories.

\head{Image-Text Contrastive Learning}
 Contrastive pretraining, as exemplified by CLIP~\cite{openai_clip,openclip} and ALIGN~\cite{jia2021scaling}, aligns global image-text embeddings within a shared latent space. Subsequent works~\cite{zhai2023siglip,li2023scaling,li2023clipa, chen2024internvl, chen2024vitamin, wang2024diffusion} focused on enhancing CLIP’s performance and improving training efficiency. 

\head{Text as Autoregressive Targets.}
Caption generation as a pretext task is another approach for visual representation learning. SimVLM~\cite{wang2022simvlm} trains encoder-decoder architectures to autoregressively predict captions, while CapPa~\cite{CapPa} trains vision encoders through sequence prediction. These methods often retain modular designs or auxiliary components like contrastive losses~\cite{yu2022coca}. 
Our work aligns with this category but removes architectural fragmentation by jointly modeling image patches and text tokens in a single Transformer. 
We find that the pre-trained Single Transformer learns transferable vision representations, enabling it to handle downstream multimodal understanding tasks and function as a vision encoder without modifications.

\section{\model: Training a Single Transformer for Vision and Language}
\label{sec:method}

\subsection{Model Architecture}

\begin{figure*}[htp]
    \centering
    \includegraphics[width=0.95\textwidth]{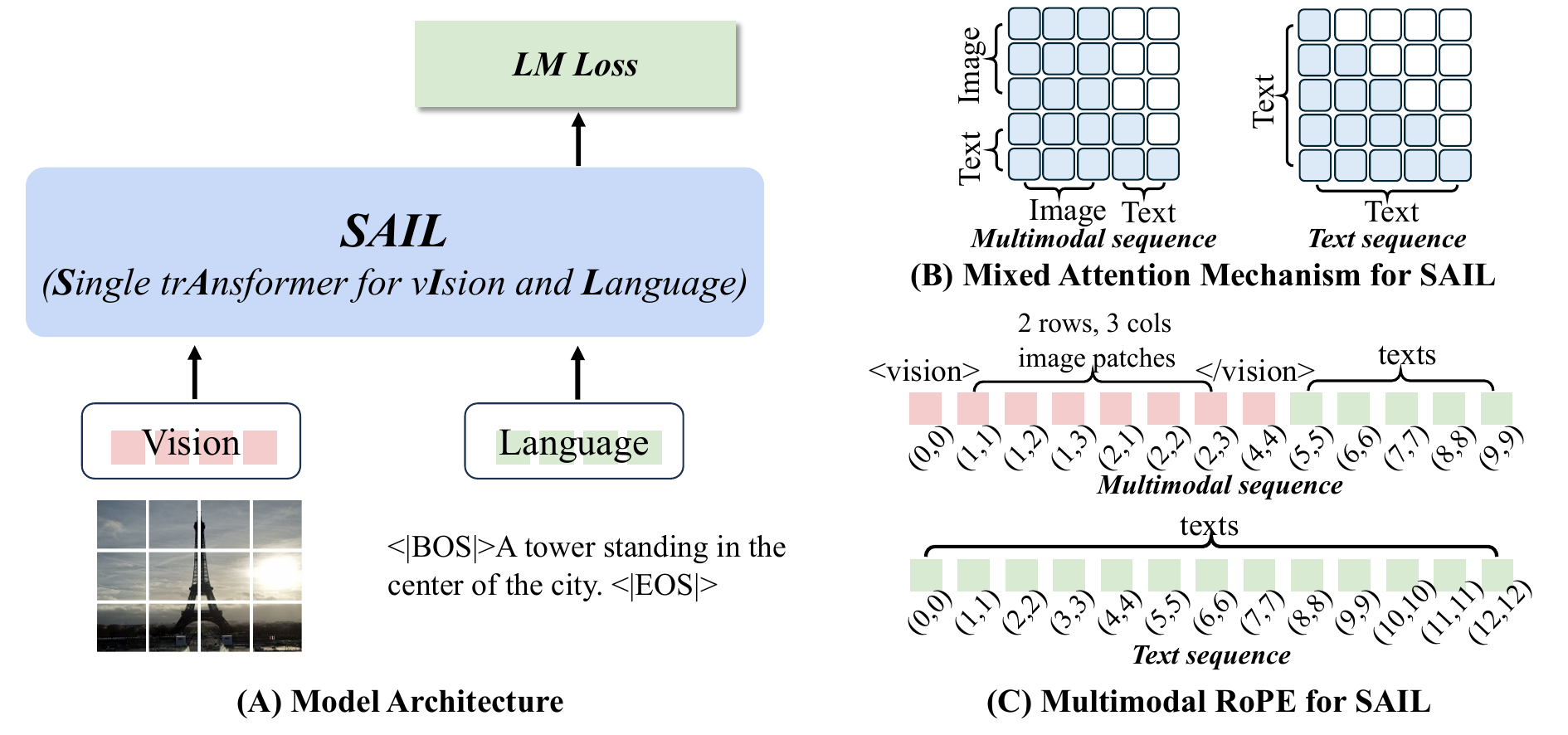}
    \caption{\textbf{Model architecture and micro-designs for \boldmodel.} 
    \textbf{(A) Model Architecture:} \model is a unified transformer that processes both images and texts without extra module designs. \textbf{(B) Mixed Attention Mechanism:} we adopt bidirectional attention for image patches from the same image and causal attention for text tokens. Examples for a multimodal sequence and a text sequence are provided. Colored squares represent ``allow to attend'' and white squares indicate ``prevent from attending''. 
    \textbf{(C) Multimodal RoPE:} an illustration of the multimodal rotary position embedding for \model, with examples for a multimodal sequence and a text sequence.}
    \label{fig:model}
    \vspace{-1em}
\end{figure*}

\model is built upon a unified Transformer architecture (Figure~\ref{fig:model}(A)) that processes multimodal inputs through streamlined, modality-specific preprocessing. 
For text, raw input is tokenized using the language model's tokenizer and then transformed into embeddings via the textual embedding module. 
For images, we partition the input into fixed-size patches and project them into continuous embeddings via a linear projection. 
Additionally, we maintain a list of special tokens explicitly designed for visual modality encoding: \texttt{<vision>} and \texttt{</vision>} tokens mark the beginning and end of an image patch span, respectively. 
In multimodal scenarios, such as image-text pairs, these embeddings are concatenated into a single sequence and fed into the Transformer, enabling joint cross-modal interactions through unified self-attention layers. 
This design eliminates the need for modality-specific encoders, which efficiently processess heterogeneous data within a single-transformer framework.

\head{Bidirectional attention within image patches.} 
While existing multimodal large language models (MLLMs)~\cite{llava,llava1.6,zhu2023minigpt,li2024llavaonevision}  predominantly adopt causal attention for autoregressive sequence modeling, our experiments reveal that enabling full bidirectional attention among tokens from the same image significantly enhances visual representation learning and boosts downstream vision-language task performance. Note that previous Single Transformer works~\cite{diao2024EVE,chen2024SOLO,luo2024mono-internvl,emu3} have only utilized causal attention, without exploring the potential of mixed attention mechanisms.

As illustrated in Figure~\ref{fig:model}(B), for \model we implement a mixed attention scheme: 
(1) For text tokens, we preserve causal attention to maintain autoregressive generation capabilities, allowing each token to attend only to its predecessors. 
(2) For image tokens, we activate full bidirectional attention within each image patch group, empowering every visual token to interact with all others in the same image. 
This design captures holistic spatial relationships and contextual dependencies among visual elements, addressing the under-explored potential of attention mechanisms in cross-modal alignment. 
The improved interaction paradigm not only refines vision-language feature fusion but also provides stronger inductive biases for complex reasoning tasks.

\head{Multimodal Rotary Position Embeddings.} 
Following~\cite{qwen2vl}, we implement Multimodal RoPE (M-RoPE) in \model to harmonize positional modeling for multimodal inputs. The method decomposes positional encoding into two axes: height, and width. For text tokens, all axes share uniform position IDs (aligned with 1D-RoPE), whereas for images, height/width IDs adaptively map to token coordinates, as is shown in Fig~\ref{fig:model}(C). 
Notably, position indexing is sequentially initialized across modalities (\eg, starting from images before extending to subsequent text), preserving inter-modal consistency. M-RoPE not only improves positional sensitivity but also constrains absolute position magnitudes for visual tokens, facilitating robust generalization to extended sequences in inference.

\subsection{Pretraining}

% \lxt{This section is not that detailed. We can list the datasets and samples that we used in one table as shown in SOLO or InternVL paper.}

% \lxt{In addition, the reviewers may attack the number of training samples with EVE, EVE-v2, SOLO. I suggest to discuss here on why our training engine is different from previous works.}

We apply a two-stage curriculum to progressively strengthen the visual perception of \model while preserving its inherent language capabilities.

\head{Stage 1: Accelerated Visual Knowledge Acquisition.}
In this stage, we pretrain \model on large-scale image-text pairs to rapidly bootstrap its visual understanding. To maximize data throughput, we uniformly resize all images to a lower resolution (\eg, $224 \times 224$), reducing multimodal sequence lengths and enabling the model to process more samples within fixed training time. To prevent catastrophic forgetting of linguistic knowledge, we interleave pure text corpora with multimodal data during training. This hybrid approach ensures efficient exposure to visual patterns while maintaining robust language proficiency.

\head{Stage 2: Enhancing Any-Resolution Image Understanding.}
Real-world applications require robustness to images of varying resolutions and aspect ratios, such as documents, charts, or infographics. Following prior works~\cite{diao2024EVE,chen2024SOLO}, we extend pretraining with an any-resolution strategy: images retain their native resolutions during processing, and positional embeddings adapt dynamically to arbitrary spatial dimensions. This stage further refines \model’s ability to model fine-grained visual details (\eg, tabular structures, text-rich graphics) while continuing to incorporate text-only data for language capability preservation.

\head{Pretraining Objective.}
Throughout both stages, we optimize the standard language modeling loss only on text tokens. Image patches and special visual tokens are excluded from loss computation. 

\subsection{Supervised Fine-tuning}
During the Supervised Fine-tuning (SFT) stage, we train \model on publicly available, multi-source instruction datasets to enhance its understanding of complex linguistic instructions and diverse dialogue patterns critical for real-world deployment. 
This phase fine-tunes the entire network architecture, focusing on aligning the model’s responses with human intent through exposure to varied instructional formats and multimodal interactions.

Table~\ref{tab:training_data_stages} shows the details of training datasets for pretraining and supervised fine-tuning (SFT) across all stages. 
During Stage 1 pretraining, we utilized mixed multimodal and pure text datasets including Recap-DataComp-1B~\cite{recap_datacomp1b} and SlimPajama~\cite{cerebras2023slimpajama}, with images at a resolution of 224x224, totaling 512M image-text pairs. 
In Stage 2, the pretraining datasets include Capfusion~\cite{yu2024capsfusion}, self-curated OCR data from LAION COCO~\cite{schuhmann2022laion5b}, InfinityMM Stage 2 subset~\cite{infinitymm}, and SlimPajama, utilizing the any resolution (AnyRes) strategy, with a combined total of 86M image-text pairs along with text data. 
The SFT stage employed the InfinityMM Stage 3 subset, processed at any resolution, containing 6M image-text pairs. 

\begin{table}[htp]
\centering\small
\setlength{\tabcolsep}{3pt}
\resizebox{\columnwidth}{!}
{
\begin{tabular}{clcc}
\toprule
Stage                                                & Dataset                                          & Img.Res                                       & Num  \\ \midrule
\multicolumn{1}{c|}{\multirow{2}{*}{Pretraining S1}} & \multicolumn{1}{l|}{Recap-DataComp-1B~\cite{recap_datacomp1b}}           & \multicolumn{1}{c|}{\multirow{2}{*}{224x224}} & 512M \\
\multicolumn{1}{c|}{}                                & \multicolumn{1}{l|}{SlimPajama~\cite{cerebras2023slimpajama}}                  & \multicolumn{1}{c|}{}                         & -    \\ \hline
\multicolumn{1}{c|}{\multirow{4}{*}{Pretraining S2}} & \multicolumn{1}{l|}{Capfusion~\cite{yu2024capsfusion}}                   & \multicolumn{1}{c|}{\multirow{4}{*}{AnyRes}}  & 60M  \\
\multicolumn{1}{c|}{}                                & \multicolumn{1}{l|}{OCR from LAION COCO~\cite{schuhmann2022laion5b}}         & \multicolumn{1}{c|}{}                         & 7M   \\
\multicolumn{1}{c|}{}                                & \multicolumn{1}{l|}{InifinityMM Stage 2 subset~\cite{infinitymm}}  & \multicolumn{1}{c|}{}                         & 19M  \\
\multicolumn{1}{c|}{}                                & \multicolumn{1}{l|}{SlimPajama~\cite{cerebras2023slimpajama}}                  & \multicolumn{1}{c|}{}                         & -    \\ \hline
\multicolumn{1}{c|}{SFT}                             & \multicolumn{1}{l|}{InifinityMM Stage3~\cite{infinitymm} } & \multicolumn{1}{c|}{AnyRes}                   & 6M  \\ \bottomrule
\end{tabular}
}
\vspace{-5pt}
\caption{
\textbf{Details of training datasets used across all stages.} 
``Img.Res'' refers to the image resolution settings applied during each training stage. All   datasets listed are publicly available. Note that these settings represent the   default configuration    for standard \model training, while separate  settings are used for scaling experiments and ablation studies.
}
\label{tab:training_data_stages}
\vspace{-1.5em}
\end{table}

\section{Experiment}
\label{sec:exp}
\begin{table*}[th]
    \centering
    \setlength{\tabcolsep}{0.075cm}
    \resizebox{\linewidth}{!}{
    \begin{tabular}{l ccc| cccccc cc cc ccc}
        \toprule
        \multirow{2}{*}{Method} & \multirow{2}{*}{\#Param} & \multirow{2}{*}{\#Data} & \multirow{2}{*}{\#Vtoken} 
        & \multicolumn{6}{c}{General VQA} &\multicolumn{2}{c}{Hallucination} & \multicolumn{2}{c}{Math\&knowledge} &\multicolumn{3}{c}{OCR VQA}\\
        \cmidrule(lr){5-10} \cmidrule(lr){11-12} \cmidrule(lr){13-14} \cmidrule(lr){15-17}
        & & & & MMS* & MMB$^\text{en}$ 
        & SEED$^\text{I}$ & MMV & MME & RWQA & POPE
        & Hallu & SQA$^\text{I}$  & MathV & TQA
        & AI2D  & OCRB \\
        \midrule
        % \rowcolor{gray!17}
        \multicolumn{4}{l|}{\emph{Modular MLLMs:}} &\multicolumn{13}{l}{}\\
        InternVL-1.5~\cite{chen2024far}
        & 2.2B &-- / -- &3328
        & 46.7 &70.9
        & 69.8 & 39.3 &\underline{1902} &57.9&\textbf{88.3}
        & \textbf{37.3} & \underline{84.9} &41.3&\underline{70.5}
        &69.8  & \textbf{654}
        \\
        QwenVL-Chat~\cite{bai2023qwen-vl}
        & 7B &7.2B / 50M &256
        & 34.5 & 60.6 
        & 58.2 & -- &1848 & 49.3 &--
        & \underline{36.8} & 68.2 & 35.3 & 61.5 
        & 45.9 & 488
        \\
        LLaVA-1.5~\cite{llava15}
        & 7B &0.4B+ / 665K &576
        & 33.1 & 64.3 
        & 64.3 & 30.5 &1859& 54.8  &85.9
        & 27.6 & 66.8 & 25.5& 46.1 
        & 54.8 & 318
        \\
        LLaVA-1.6~\cite{llava1.6}
        &7B &0.4B+ / 760K &2880
        & 37.6 & 67.4 
        & 64.7 & \underline{43.9} &1842 & 57.8 &\underline{86.4}
        & 27.6 & 70.2& 32.5 & 64.9 
        & 66.6 & \underline{532} 
        \\
        Cambrian-1~\cite{tong2024cambrian}
        &8B &10B+ / 7M &576
        &\underline{50.7} &\underline{75.9} 
        &\underline{74.7} &--  &--&\underline{64.2} &--
        & 30.6 &80.4  & \underline{48.1}&\textbf{71.7}
        &\underline{73.0}  &--
        \\
        LLaVA-OneVision~\cite{li2024llavaonevision} 
        &7B &10B+ / 3.2M &7290
        &\textbf{60.9} &\textbf{81.7}  
        &\textbf{74.8} &\textbf{58.8} &\textbf{1998}&\textbf{65.5}  &--
        &-- &\textbf{96.6} &\textbf{56.1}&-- 
        &\textbf{81.6} &--
        \\
        \midrule
        % \rowcolor{gray!17}
        \multicolumn{4}{l|}{\emph{Single Transformer-based MLLMs:}}  &\multicolumn{13}{l}{}\\
        Fuyu~\cite{fuyu-8b} 
        & 8B &-- / -- &--
        & 34.4  & 10.7 
        & 59.3   & 21.4 &-- & 43.7  &84
        & 29.8   & 56.8& 30.2   & -- 
        & 46.8& 366
        \\
        Chameleon~\cite{chameleon}
        &7B &1.4B+ / 1.8M &1024
        &31.1 &31.1   
        &30.6 & 8.3 &170 & 39 &19.4
        & 17.1  & 47.2  &22.5&4.8
        &46.0 & 7.0 
        \\
        EVE~\cite{diao2024EVE}
        &7B &33M / 1.8M &2304
        & -- & 52.3
        & 64.6 & 25.7 & 1628& --  & 85.0  
        & -- &64.9 &  -- & 56.8 
        & 61.0 & 398
        \\
        SOLO~\cite{chen2024SOLO}
        & 8B &43.7M / 2M &1024
        & 35.8  &  67.7
        & 64.4 & 30.4 & 1260& 44.7 &78.6
        & 40.4   &73.3    & 32.9  & 25.0
        & 61.4  & 126  
        \\
        Mono-InternVL~\cite{luo2024mono-internvl}
        & 3B &1.3B / 7M &6400
        & --  & 65.5
        & 67.4  & 40.1 & \textbf{1875}& -- &--
        & \underline{45.7} &\underline{93.6}   & \underline{45.7} &\underline{72.6}
        & 68.6  & \underline{767}
        \\
        Emu3~\cite{emu3}
        & 8B &-- / -- &16K
        & \underline{46.6} & 58.5
        & 68.2 & 37.2 & --& 57.4 & 85.2
        & 31.7 & 89.2  & 31.3& 64.7
        & 70.0  & 687 
        \\
        EVE2~\cite{diao2025EVEv2}
        & 7B &92M / 7.3M &2500
        &-- & \underline{66.3}
        & \underline{71.4} & \underline{45.0} & 1709& \underline{62.4} & \textbf{87.6}
        &-- & \textbf{96.2} &-- &71.1
        & \underline{74.8}  &702
        \\
        \textbf{\model}
        & 7B & 600M / 6M & 3600
        &\textbf{53.1} &\textbf{70.1}
        &\textbf{72.9} &\textbf{46.3} & \underline{1719}&\textbf{63.9} & \underline{85.8}
        &\textbf{54.2} &93.3 &\textbf{57.0} &\textbf{77.1}&\textbf{76.7}  &\textbf{783} 
        \\
        \bottomrule
        \\
        \end{tabular}
        }
    \vspace{-15pt}
    \caption{Comparison with existing vision-language models on various vision-language benchmarks, 
    including MMS*: MMStar~\cite{mmstar};
    MMB$^\text{en}$: MMBench-EN~\cite{liu2024mmbench};
    SEED$^\text{I}$: SEEDBench-Img~\cite{seedbench};
    MMV: MMVet~\cite{mmvet};
    MME~\cite{mme};
    POPE~\cite{POPE}; Hallu: HallusionBench~\cite{hallusionbench};
    SQA$^\text{I}$: ScienceQA-Img~\cite{scienceqa}; 
    TVQA: TextVQA~\cite{textvqa};   
    MathV: MathVista$_{\text{MINI}}$~\cite{lu2023mathvista};
    AI2D~\cite{AI2D};
    RWQA: RealWorldQA~\cite{VLM:Grok-1.5};
    OCRB: OCRBench~\cite{liu2024ocrbench}.
    Note that \#A-Param denotes the number of activated parameters;
    \#Data represents the pre-training / fine-tuning data volume;
    \#Vtoken indicates the maximum image patch tokens.
    For MME, we report the sum of  perception and cognition scores.
    The top two results are highlighted in \textbf{bold} and \underline{underline}, respectively. 
All results are derived from those reported in other papers and the official reproduction results from the OpenCompass leaderboard~\cite{duan2024vlmevalkit}. Our results are obtained by VLMEvalKit~\cite{duan2024vlmevalkit}.
    }
    \label{tab:main_table_multimodal_benchmark}
    \vspace{-4mm}
\end{table*}

\subsection{Experimental Settings}

\head{Evaluation Benchmarks.}
For evaluation of vision and language tasks, we evaluate \model and existing MLLMs on a broad range of multimodal benchmarks. Specifically, MLLM benchmarks encompass MMBench-EN~\cite{liu2024mmbench}, SEEDBench-IMG~\cite{seedbench}, MMVet~\cite{mmvet}, 
MME~\cite{mme}, HallusionBench~\cite{hallusionbench},
MathVista$_{\text{MINI}}$~\cite{lu2023mathvista}, and
OCRBench~\cite{liu2024ocrbench}. 
Visual question answering benchmarks include TextVQA~\cite{textvqa}, ScienceQA-IMG~\cite{scienceqa}, AI2D~\cite{AI2D}, MMStar~\cite{mmstar}, RealWorldQA~\cite{VLM:Grok-1.5}.
 
For evaluation of vision representation learning, we conduct experiments on ImageNet-1K~\cite{deng2009imagenet} for image classification, ADE20K for semantic segmentation~\cite{zhou2017ade20k}, and ARO~\cite{yuksekgonul2022aro} for attribute, relation, and ordering.

\head{Implementation Details.} 
For pretraining, we initialize \model from the Mistral-7B-v0.1 base LLM and set the patch size to 14. We modify Megatron~\cite{shoeybi2019megatron} to support \model's multimodal input. 
Pretraining uses 128 NVIDIA A100 80G GPUs with 2-way tensor parallelism and 64-way data parallelism. 
The learning rate is set at 5e-5 and decays cosinely to a minimum of 5e-6. 
For training efficiency, we concatenate sequences from different data samples into one long sequence of 32,768 tokens, adjusting the attention mask to ensure that tokens from different samples do not attend to each other. 
We use a round-robin approach to interleave image-text packed sequences and pure text packed sequences, configuring the global batch to contain approximately 16K image-text pairs.

For Supervised Fine-Tuning (SFT), the global batch size is set to 512. Training is performed for one epoch with a maximum learning rate of 1e-5, following a linear warm-up phase and then transitioning to a cosine decay schedule.

%%%%%%% Haochen
For vision, we load the checkpoint after Stage 1 pretraining and keep it \textit{frozen} for downstream evaluations, including (1) image classification on ImageNet-1K~\cite{deng2009imagenet}, (2) semantic segmentation on ADE20K~\cite{zhou2017ade20k}, and (3) attribute, relation and,
ordering on the ARO benchmark~\cite{yuksekgonul2022aro}.
Specifically, (1) for image classification, we utilize an attention-based classifier~\cite{el2024aim} with 90 epochs of linear probing, where detailed configurations are mostly obtained from common practices~\cite{he2022masked, wang2023bootstrap, wang2023hard}.
Images are resized to 224$\times$224 and the global batch size is 8,192 across 8 A100 (80G) GPUs.
(2) For semantic segmentation, we adopt ViT-Adapter~\cite{chen2022vitadapter} with UperNet~\cite{xiao2018upernet} as the segmentation decoder.
The implementation is based on MMSegmentation~\cite{mmseg2020} with 80k training iterations.
The input resolution is 512$\times$512 and the global batch size is 16 across 8 A100 (80G) GPUs.
(3) For attribute, relation and,
ordering, we regard the negative of the caption loss over each image-text pair as the similarity metric for retrieval.

\subsection{Experimental Results}
\subsubsection{Results on Vision Language Tasks}
As shown in Table~\ref{tab:main_table_multimodal_benchmark}, we compare \model against existing MLLMs across 13 vision-language benchmarks. \model consistently outperforms other Single Transformer-based models like Fuyu~\cite{fuyu-8b}, EVE~\cite{diao2024EVE}, SOLO~\cite{chen2024SOLO}, Mono-InternVL~\cite{luo2024mono-internvl}, and EVE2~\cite{diao2025EVEv2} across diverse vision-language tasks. 
This demonstrates that \model can achieve significant performance gains and push the boundaries of Single Transformer-based MLLMs without needing extra component designs or auxiliary training losses.
Moreover, when compared to methods employing discrete vision tokens (e.g., Chameleon and Emu3), \model demonstrates superior performance. These results validate that scaling up single-transformer pretraining effectively enhances cross-modal alignment between images and text.
Compared to the state-of-the-art modular MLLM LLaVA-OneVision~\cite{li2024llavaonevision}, \model achieves comparable performance on some benchmarks, such as MMStar, SEEDBench-IMG, and RealWorldQA. 
While the performance of Single Transformer-based MLLMs still lags behind modular MLLMs in certain areas, we hypothesize that scaling the pretraining data volume or incorporating higher-quality instruction-tuning data will bridge the remaining performance gap.

%%%%% Haochen
\subsubsection{Results on Vision Representation Learning}

In this section, we compare the quality of learned visual representations of our \model with other Single Transformer-based alternatives, including EVE~\cite{diao2024EVE}, EVE2~\cite{diao2025EVEv2}, and SOLO~\cite{chen2024SOLO}.

\begin{table}[t]
    \centering\small
    \setlength{\tabcolsep}{7.6pt}
    \begin{tabular}{l cc ccc}
    \toprule
    \multirow{2}{*}{Method} & \multicolumn{2}{c}{Classification} & \multicolumn{3}{c}{Segmentation} \\
    \cmidrule(lr){2-3}
    \cmidrule(lr){4-6}
    & Top-1 & Top-5 & mIoU & mAcc & aAcc \\
    \midrule
    EVE~\cite{diao2024EVE} & 42.03 & 65.77 & 27.12 & 35.89 & 72.91 \\
    EVE2~\cite{diao2025EVEv2} & 44.86 & 69.41 & 40.85 & 53.53 & 79.31 \\
    SOLO~\cite{chen2024SOLO} & 59.10 & 80.89 & 35.11 & 44.81 & 76.02 \\
    \textbf{\model} & \textbf{84.95} & \textbf{97.59} & \textbf{55.30} & \textbf{67.24} & \textbf{84.87} \\
    \bottomrule
    \end{tabular}
    \vspace{-5pt}
    \caption{
    \textbf{Comparison on image classification and semantic segmentation} with other encoder-free approaches.
    Our \model outperforms other alternatives by a large margin.
    }
    \label{tab:exp_classifcation}
\end{table}

\begin{table}[t]
    \centering\small
    \setlength{\tabcolsep}{3pt}
    \begin{tabular}{l cc cl}
    \toprule
    Method & \#Data & \#Param & ImageNet-1K & ADE20K \\
    \midrule
    OpenCLIP-H~\cite{openclip} & 2B & 0.6B & 84.4 & -- \\
    OpenCLIP-G~\cite{openclip} & 2B & 1.8B & 86.2 & 39.3$^{\dag}$ \\
    ViT-22B~\cite{dehghani2023scaling} & 3B & 22B & \textbf{89.5} & 55.3 \\
    InternViT~\cite{chen2024internvl} & 6B & 6B & 88.2 & \textbf{58.7} \\
    \textbf{\model} & \textbf{0.5B} & 7B & 85.0 & 55.3 \\
    \bottomrule
    \end{tabular}
    \vspace{-5pt}
    \caption{
    \textbf{Comparison on image classification and semantic segmentation} with other vision backbones.
    $\dag$ indicates training with head tuning using UperNet~\cite{xiao2018upernet}, while others are based on ViT-Adapter~\cite{chen2022vitadapter}.
    \model, with significantly less training data, achieves competitive performance.
    }
    \label{tab:exp_vit}
    \vspace{-1em}
\end{table}

\begin{table}[t]
    \centering\small
    \setlength{\tabcolsep}{4.5pt}
    \begin{tabular}{l cccc}
    \toprule
    \multirow{2}{*}{Method} & \multirow{2}{*}{Relation} & \multirow{2}{*}{Attribute} & \multicolumn{2}{c}{Order} \\
    \cmidrule(lr){4-5}
    & & & COCO & Flickr30K \\
    \midrule
    OpenCLIP-H~\cite{openclip} & 49.9 &	64.6 & 32.6 & 40.4 \\
    OpenCLIP-G~\cite{openclip} & 49.9 & 65.6 & 33.0 & 38.3 \\
    CLIP-B/32~\cite{openai_clip} & 59.2 & 62.9 & 48.1 & 57.9 \\
    CLIP-L/14~\cite{openai_clip} & 61.2 & 61.7 & 46.8 & 56.8 \\
    InternViT~\cite{chen2024internvl} & 59.6 & 66.0 & 73.4 & 76.3 \\
    NegCLIP~\cite{yuksekgonul2022aro} & 81.0 & 71.0 & 86.0 & 91.0 \\
    CapPa~\cite{tschannen2023cappa} & 86.7 & 85.7 & 98.8 & 99.2 \\
    \textbf{\model} & \textbf{100.0} & \textbf{99.5} & \textbf{100.0} & \textbf{100.0} \\
    \bottomrule
    \end{tabular}
    \vspace{-5pt}
    \caption{
    \textbf{Comparison on attribute, relation, and ordering (ARO)} with other vision backbones.
    \model almost encodes compositional relationships between objects and attributes perfectly.
    }
    \label{tab:exp_aro}
    \vspace{-1em}
\end{table}

\head{Classification and Segmentation.}
As demonstrated in Table~\ref{tab:exp_classifcation}, our method, \model, achieves a Top-1 accuracy of 84.95\% and a Top-5 accuracy of 97.59\% on the validation set of ImageNet-1K~\cite{deng2009imagenet}, significantly outperforming state-of-the-art alternatives~\cite{diao2024EVE, diao2025EVEv2, chen2024SOLO}. 
In the segmentation task, \model also demonstrates superior performance with an mIoU of 55.30\%, an mAcc of 67.24\%, and an aAcc of 84.87\%, illustrated in Table~\ref{tab:exp_classifcation}.
These results indicate that \model is effective in both classification and segmentation tasks, offering substantial improvements over existing methods.
In Table~\ref{tab:exp_vit}, even when comparing with other state-of-the-art vision backbones, our \model manages to achieve remarkable competitive performance with significantly less training data, demonstrating the scaling property of \model.

\head{Attribute, Relation, and Ordering.}
To systematically evaluate the ability of \model to understand different types of relationships, attributes, and order information, we conduct experiments on the ARO benchmark~\cite{yuksekgonul2022aro}.
As demonstrated in \Cref{tab:exp_aro}, \model encodes compositional relationships between objects and attributes almost perfectly, significantly surpassing other state-of-the-art vision backbones.

For additional vision-related tasks, please refer to Pixel-\model~\cite{zhangta2025pixelsail} for \model's downstream capabilities in pixel-grounded understanding.

\subsection{Properties of Single Transformer}
\subsubsection{Scaling Properties.}
\head{Model Scaling:} We selected models of different sizes: \model-0.5B, \model-3B, and \model-7B (\model by default) for our experiments. Each model underwent Stage 1 pretraining on a mixed multimodal and pure text dataset, encountering 512M image-text pairs. Subsequently, they were fine-tuned on the LLaVA-mix-665K dataset using the any resolution (anyres) strategy. 
We evaluated the models based on their performance on vision and language benchmarks after supervised fine-tuning.

\begin{figure}[t]
    \centering
    \includegraphics[width=1.0\columnwidth]{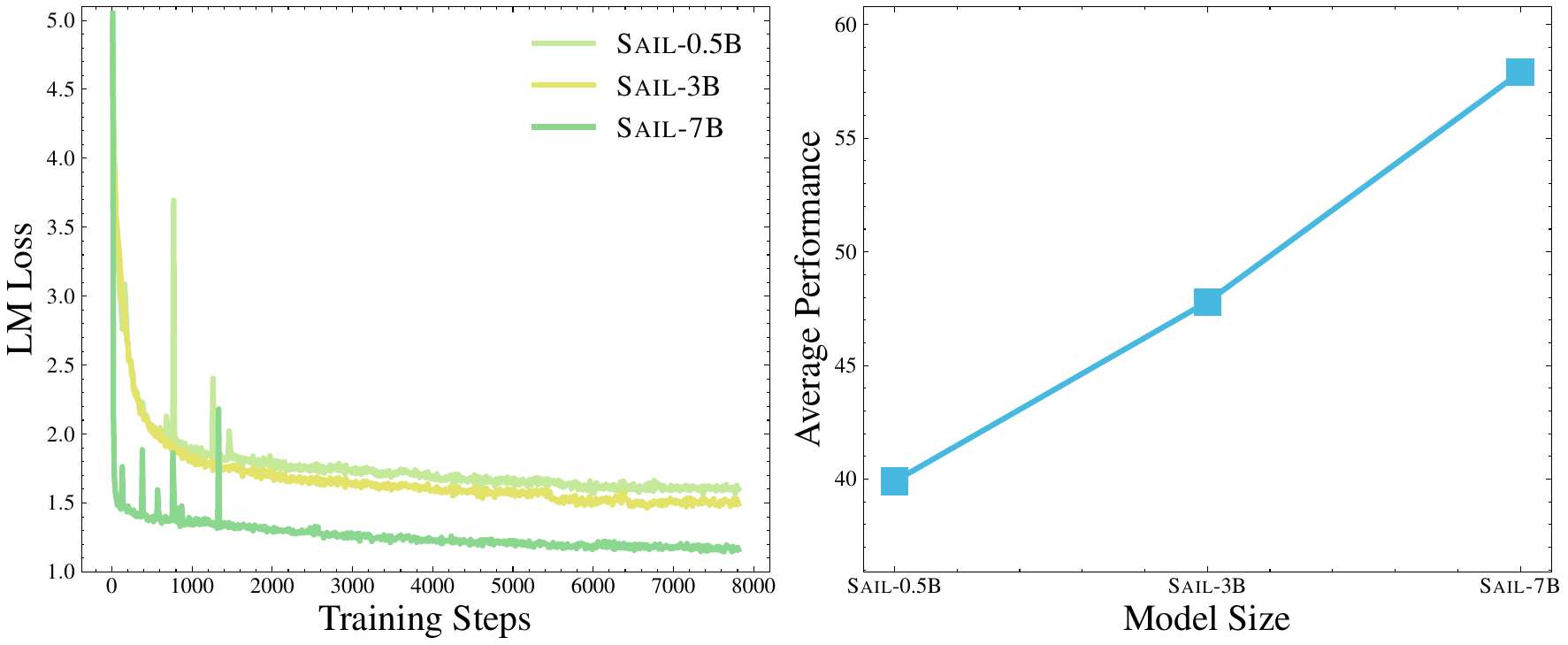}
\caption{
    \textbf{Model scaling of \boldmodel.} Left: As the model size increases, the training language modeling loss gradually decreases. Right: As the model size increases, performance on downstream VLM tasks progressively improves.
    }

    \label{fig:exp_modeling_scaling}
    \vspace{-.5em}
\end{figure}

The normalized performance of \model against model size is plotted in Figure~\ref{fig:exp_modeling_scaling}. As the model size scales up, we observe a corresponding enhancement in performance. Additionally, as shown on the left side of Figure~\ref{fig:exp_modeling_scaling}, the training language modeling loss decreases with increasing model size. This reduction in training loss indicates that larger models have a greater capacity to learn multimodal alignments effectively, enabling them to capture complex relationships between vision and language more accurately. The improved learning capacity directly translates to better performance on downstream VLM tasks, showcasing the benefits of scaling up the Single Transformer architecture.

\head{Data Scaling:} we compared \model with its modular MLLM counterpart. For the modular MLLM, we used SigLIP-SO~\cite{zhai2023siglip} as the vision encoder, and the language model shared the same architecture and initialization parameters as \model. Both models were pre-trained using Pretraining stage-1 setting, with \model encountering 32M, 128M, and 512M image-text pairs during training, followed by fine-tuning on the LLaVA-mix-665K dataset.  \textit{All parameters} of both models are trainable. Both models employ an identical number of input tokens for images and text.
The normalized performance of both models is plotted in Figure~\ref{fig:teaser}(A). The results show that in the low-data regime (32M), \model's performance lags behind the modular MLLM, likely due to SigLIP's prior training on 40B samples. 
However, as the data scales, \model exhibits a steeper performance curve, indicating more promising data scaling properties. 
At 512M image-text pairs, \model achieves performance comparable to the modular MLLM in our evaluation subset. This demonstrates the single transformer's superior data scalability, even without a pretrained vision encoder. 

Quantitative results on evaluated benchmark tasks of model scaling and data scaling are tabulated in the appendix.

\begin{figure}[t]
    \centering
    \includegraphics[width=1.0\columnwidth]{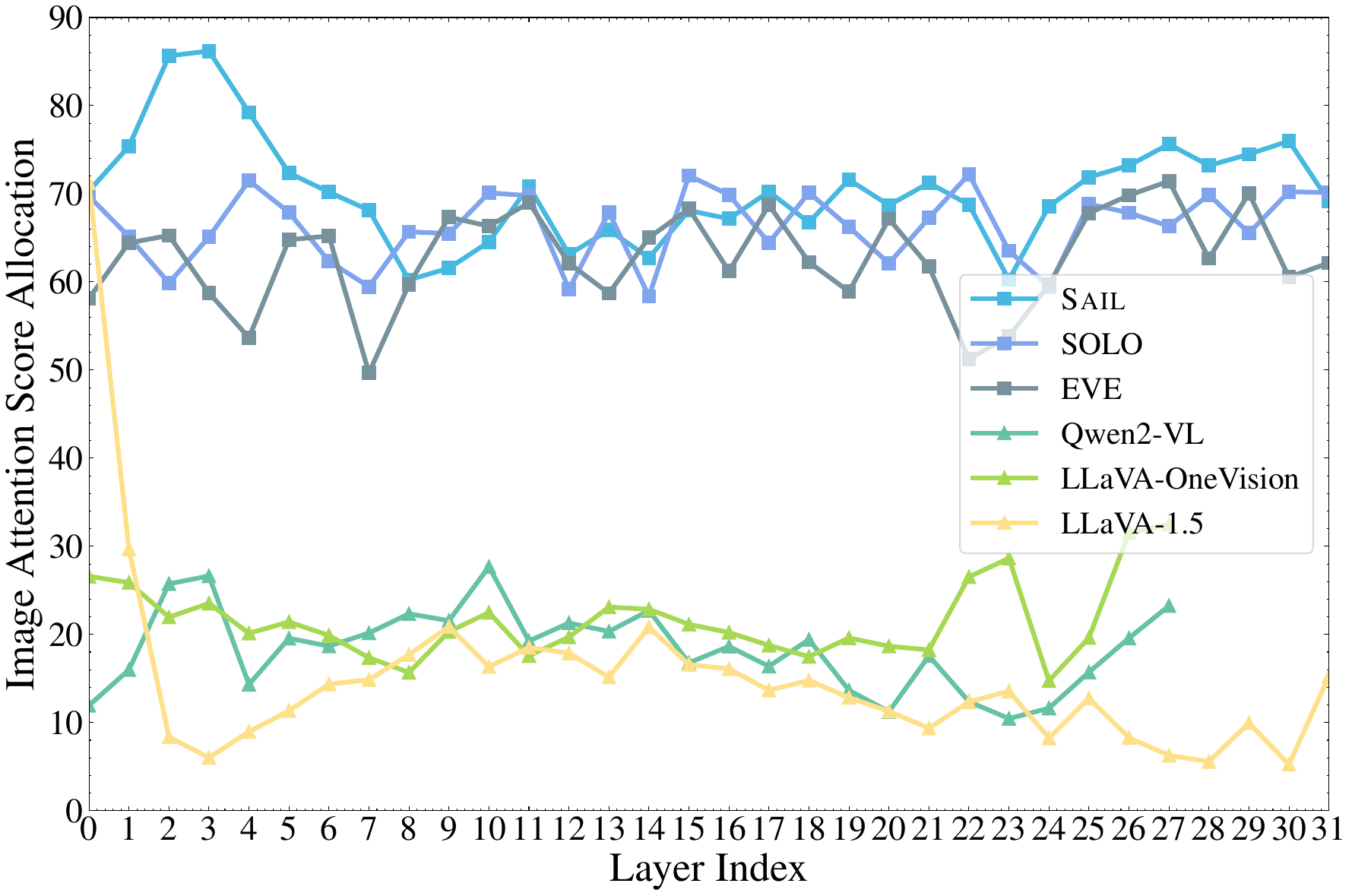}
    \caption{
      \textbf{Image Attention Score Allocation}: The figure shows the proportion of image attention scores across different transformer layers for Single Transformer-based MLLM and modular MLLM when predicting tokens. Single Transformer-based MLLM generally allocates higher attention weights to image tokens compared to modular MLLM.
    }
    \label{fig:exp_attn_alloc_by_layer}
    \vspace{-0.5em}
\end{figure}

\subsubsection{Information Flow Pattern}
\head{Different attention pattern compared to modular MLLM:}
since our comparative experiments show that the Single Transformer model exhibits more promising data scaling properties, we conducted an analysis of the trained \model model and its modular MLLM counterpart. Specifically, we followed the methodology from FastV~\cite{fastv} to analyze the attention score distribution for each predicted token given an image and a user query. This analysis focuses on how much attention is allocated to image tokens during token prediction.
We selected 1000 samples from various datasets including VQAv2, GQA, TextVQA, DocVQA, MME, SEEDBench-IMG, MMBench, and some self-curated dialog examples. For each model prediction, we computed the average attention scores assigned to previous image tokens by the output token.

We conducted a comparative experiment between Single Transformer-based MLLMs and modular MLLMs. The Single Transformer-based MLLMs included \model, SOLO~\cite{chen2024SOLO}, and EVE~\cite{diao2024EVE}, while the modular MLLMs included Qwen2-VL~\cite{qwen2vl}, LLaVA-OneVision~\cite{li2024llavaonevision}, and LLaVA1.5~\cite{llava15}.

The results are depicted in Figure~\ref{fig:exp_attn_alloc_by_layer}. Single Transformer-based MLLMs allocate between 60\% and 80\% of attention scores to image tokens across all layers when predicting tokens. In contrast, modular MLLMs such as Qwen2-VL and LLaVA-OneVision allocate only 10\% to 30\% of attention scores to image tokens across different layers. For LLaVA1.5, which does not update the ViT parameters during supervised fine-tuning (SFT), the image attention score is relatively high in the first two transformer layers but declines sharply in subsequent layers.

From this experiment, we can conclude that Single Transformer-based MLLMs tend to allocate a significant portion of attention to previous image tokens during prediction. In contrast, modular MLLMs allocate a smaller portion of their attention directly to image tokens, indicating a less image-centric approach in their prediction mechanism.

These findings indicate that the Single Transformer model places more emphasis on grounding its predictions in the visual information. As the model undergoes data scaling, it allocates more effective computation to image tokens, thereby enhancing its capability as a vision-centric model.

In summary, the attention pattern analysis underscores the Single Transformer’s ability to robustly integrate visual context, enabling it to scale efficiently and potentially outperform modular MLLMs in vision-language tasks.

\subsubsection{Task-Specific Performance Analysis}
We dissect \model’s strengths and limitations through targeted case studies:

\head{Strengths: Spatial Reasoning.} 
\model excels at tasks requiring precise spatial location. As shown in Table~\ref{tab:strength_and_weakness}, under the setting of our data scaling experiment, it outperforms the modular counterpart by 61.7 points on the MME Position split and 21.8\% on MMBench Physical Relation questions. The unified architecture likely enables tighter coupling between visual geometry and linguistic descriptions.

\head{Weaknesses: World Knowledge.} 
Conversely, \model falls short in tasks that demand extensive world knowledge. As shown in Table~\ref{tab:strength_and_weakness} \model underperforms in the MME celebrity and art splits compared to the modular MLLM. This underperformance can be attributed to \model's lack of diverse domain-specific data during pretraining, a gap that was not sufficiently addressed during supervised fine-tuning. Modular MLLMs, with their pretrained vision encoders like CLIP~\cite{openai_clip,openclip} or SigLIP~\cite{zhai2023siglip}, have a broader knowledge base and therefore handle such tasks more effectively. We hypothesize that scaling up \model’s pretraining data diversity could help bridge this gap, enhancing its performance on knowledge-intensive tasks.

\begin{table}[t]
\setlength{\tabcolsep}{2pt}
\resizebox{\columnwidth}{!}
{
\begin{tabular}{cccccc}
\toprule
\multirow{2}{*}{Method}             & \multicolumn{2}{c}{MMBench~\cite{liu2024mmbench}}                                                                                                 & \multicolumn{3}{c}{MME~\cite{mme}}        \\     \cmidrule(lr){2-3}
\cmidrule(lr){4-6}
             & \begin{tabular}[c]{@{}c@{}}Physical\\ Relation\end{tabular} & \begin{tabular}[c]{@{}c@{}}Celebrity \\ Relation\end{tabular} & Position & Posters & Celebrity \\ \midrule
Modular MLLM & 30.4                                                        & 50.5                                                          & 98.3     & 134.0   & 100.3     \\
\model         & 52.2                                                        & 88.9                                                          & 160.0    & 108.2   & 75.0      \\ 
\bottomrule
\end{tabular}
}
\vspace{-0.5em}
\caption{
\textbf{Performance Comparison of \model and Modular MLLM in MMBench and MME Tasks}: the strengths of \model in spatial reasoning tasks (MMBench Physical Relation and MME Position split) and its weaknesses in world knowledge tasks (MMBench Celebrity Relation and MME Celebrity and Posters splits).
}
\label{tab:strength_and_weakness}
\vspace{-.5em}
\end{table}

\subsection{Empirical Observations on Basic Factors}
To guide scalable training of single-transformer MLLMs, we conduct ablation studies on two critical design choices using \model-0.5B pretrained on 128M image-text pairs and fine-tuned on LLaVA-mix-665K. 
Performance is evaluated through zero-shot image classification after pretraining~\cite{CapPa} and vision-language benchmarks after SFT.

\head{Bidirectional Attention for Image Patches with Multimodal Position Encoding.} 
We compare two approaches for integrating image patches into the transformer: (1) Causal attention with 1D positional encoding, using a \texttt{<vision\_sep>} token to demarcate image rows. (2) Full bidirectional attention for image patches paired with multimodal rotary position embeddings (RoPE), which jointly encode spatial coordinates (\eg, 2D grid positions) and text token positions. 
As shown in Table~\ref{tab:exp_ablation}, configuration of using bidirectional attention with multimodal RoPE significantly improves performance on vision-language tasks, with a particularly notable gain of 3.1\% on TextVQA. This suggests that enabling cross-patch interactions during pretraining enhances visual representation learning and tightens cross-modal alignment.

\head{Interleaving Pure Text Data During Pretraining.} We analyze the impact of mixing SlimPajama text data with image-text pairs during pretraining. The results, as presented in Table~\ref{tab:exp_ablation} \#2, reveal that mixing in pure text data consistently improves performance across vision and language benchmarks. This finding underscores the importance of preserving language capabilities in the LLM when training Single Transformer models, as maintaining strong language skills is crucial for building a multimodal model capable of complex reasoning. Currently, incorporating text data in training is one of the effective methods to maintain the language abilities of the model.

In conclusion, our ablation studies identify key design choices for training \model effectively. Using bi-directional attention with multimodal rotary position embeddings enhances visual perception, while incorporating pure text data preserves essential language capabilities for robust multimodal performance.

\begin{table}[t]
\centering
\resizebox{\linewidth}{!}{
\setlength{\tabcolsep}{3pt}
\begin{tabular}{lccccccc}
\toprule
\bf Exp. Setting               & \bf VQAv2 & \bf GQA & \bf SQA & \bf TQA  & \bf SEED-I \\ \Xhline{2\arrayrulewidth}
\textit{Default}  &  59.1  & 46.9   &  59.6 & 20.1 & 35.1   \\
\#1 \textit{No Img full attn}   &  57.8  & 45.2  & 58.7 & 16.2  & 33.8  \\
\#2 \textit{No pure text in PT} & 56.3 & 42.1 & 48.6 & 18.3 & 32.4 \\
\bottomrule
\end{tabular}
}
\vspace{-0.5em}
\caption{
\textbf{Ablation Study on Basic Factors for \model}: This table presents the impact of different ablation settings on the performance of \model across VQAv2~\cite{vqav2}, GQA~\cite{hudson2019gqa}, SQA~\cite{scienceqa}, TQA~\cite{textvqa}, and SEED-I~\cite{seedbench}. The default setting includes image full attention and the inclusion of pure text data in pretraining. Ablation \#1 removes image full attention, and ablation \#2 excludes pure text in pretraining.
}
\label{tab:exp_ablation}
\vspace{-.5em}
\end{table}

\vspace{-2mm}
\section{Conclusion}
\label{sec:conclusion}

In this work, we conducted an extensive analysis of Single Transformer-based MLLMs compared to modular MLLMs. 
Our investigation explored the unique properties of Single Transformers, including scalability, cross-modal information flow patterns, and visual representation capabilities. 
A series of experiments on our trained \model model demonstrated that this unified architecture achieves performance on vision-language benchmarks comparable to modular MLLMs while also functioning effectively as a vision backbone. 
Our findings highlight several advantages of Single Transformer architectures, such as superior data scalability, vision-centric information flow, and inherent capabilities as a powerful vision encoder. 
We hope our empirical findings will inspire further research to refine and enhance Single Transformer architectures, advancing the field of multimodal intelligence.

% \clearpage
% \setcounter{page}{1}
% \maketitlesupplementary

\begin{table*}[h!]
\setlength{\tabcolsep}{2pt}
\resizebox{\textwidth}{!}
{
\begin{tabular}{lll|cc|cc|cc}
\hline
\multirow{2}{*}{Exp} & \multirow{2}{*}{Model} & \multicolumn{1}{c|}{\multirow{2}{*}{LLM}} & \multicolumn{2}{c|}{Stage 1}                                 & \multicolumn{2}{c|}{Stage 2}         & \multicolumn{2}{c}{SFT}       \\ \cline{4-9} 
                     &                        & \multicolumn{1}{c|}{}                     & Data                                          & LR           & Data                  & LR           & Data              & LR        \\ \hline
Figure 1(A)          & \model, point 32M        & Mistral-7B-v0.1                           & Standard Stage 1 Data (32M image-text pairs)  & (5e-5, 5e-6) & -                     & -            & LLaVA-mix-665K    & (1e-5,0)  \\
Figure 1(A)          & \model, point 128M       & Mistral-7B-v0.1                           & Standard Stage 1 Data (128M image-text pairs) & (5e-5, 5e-6) & -                     & -            & LLaVA-mix-665K    & (1e-5,0)  \\
Figure 1(A), Table 6 & \model, point512M        & Mistral-7B-v0.1                           & Standard Stage 1 Data (512M image-text pairs) & (5e-5, 5e-6) & -                     & -            & LLaVA-mix-665K    & (1e-5,0)  \\
Figure 1(B), Table 2 & \model                   & Mistral-7B-v0.1                           & Standard Stage 1 Data (512M image-text pairs) & (5e-5, 5e-6) & Standard Stage 2 Data & (1e-5, 5e-6) & Standard SFT Data & (1e-5, 0) \\
Table 3, 4, 5        & \model                   & Mistral-7B-v0.1                           & Standard Stage 1 Data (512M image-text pairs) & (5e-5, 5e-6) & -                     & -            & -                 & -         \\
Figure 3, Table 7    & \model-0.5B              & Qwen2.5-0.5B                              & Standard Stage 1 Data (128M image-text pairs) & (5e-4, 5e-6) & -                     & -            & LLaVA-mix-665K    & (1e-5, 0) \\
Figure 3             & \model-3B                & Qwen2.5-3B                                & Standard Stage 1 Data (128M image-text pairs) & (1e-4, 5e-6) & -                     & -            & LLaVA-mix-665K    & (1e-5, 0) \\
Figure 3             & \model-7B                   & Mistral-7B-v0.1                           & Standard Stage 1 Data (128M image-text pairs) & (5e-5, 5e-6) & -                     & -            & LLaVA-mix-665K    & (1e-5,0)  \\ \hline
\end{tabular}
}
\caption{
\textbf{Experimental Configurations for Various Setups.}  The table lists the models used, the specific LLM variants, the datasets, and learning rates (LR) applied during each training stage (Pretraining Stage 1, Pretraining Stage 2, and SFT). ``Standard Stage 1 Data'', ``Standard Stage 2 Data'' and ``Standard SFT Data'' are listed in Table~\ref{tab:training_data_stages}. Specific points and tables/figures referred to in the text are also indicated.
}
\label{tab:supp_exp_config}
\end{table*}
\begin{table*}[h!]
    \centering
    \setlength{\tabcolsep}{2pt}
    \resizebox{0.85\textwidth}{!}
    {
\begin{tabular}{l|cccccccc|c}
\toprule
Model                        & VQAv2 & GQA   & SciQA-IMG & TextVQA & POPE  & MME    & MMBench & SEEDBench-IMG & Norm.avg \\ \midrule
Figure 1, modular MLLM, 32M  & 76.96 & 58.7  & 68.48     & 58.68   & 88.17 & 1599   & 69.44   & 70.31         & 61.41    \\
Figure 1, modular MLLM, 128M & 78.47 & 59.78 & 70.05     & 59.82   & 86.78 & 1638   & 68.57   & 68.11         & 61.52    \\
Figure 1, modular MLLM, 512M & 80.06 & 62.38 & 70.34     & 57.85   & 83.14 & 1379   & 70.82   & 69.83         & 61.86    \\
Figure 1, \model, 32M          & 70.51 & 57.95 & 63.32     & 31.67   & 81.77 & 1421   & 48.22   & 61.51         & 51.93    \\
Figure 1, \model, 128M         & 76.36 & 60.93 & 62.61     & 56.86   & 85.5  & 1458   & 53.94   & 66.60         & 57.91    \\
Figure 1, \model, 512M         & 78.51 & 62.06 & 67.48     & 63.94   & 86.04 & 1530   & 56.71   & 68.83         & 60.51    \\
Figure 3, \model-3B            & 67.3  & 53.2  & 63.8      & 30.9    & 66.9  & 820.8  & 44.6    & 55.4          & 47.80    \\
Figure 3, \model-0.5B          & 59.1  & 46.9  & 59.6      & 20.1    & 59.8  & 761.45 & 38.5    & 35.1          & 39.92    \\ \bottomrule
\end{tabular}
}
\caption{
 Detailed experimental results in the main paper.
}
\label{tab:supp_detailed_number}
\end{table*}
\begin{table*}[h!]
    \centering
    \setlength{\tabcolsep}{2pt}
    \resizebox{0.85\textwidth}{!}
    {
        \begin{tabular}{lcc|ccccccccccc|c}
        \toprule
        Method & Pretrain & SFT & VQAv2 & GQA & SciQA-IMG & TextVQA & POPE & MMBench & SEEDBench & DocVQA & ChartQA & AI2D & MMStar & avg \\
        \midrule
        LLaVA-1.5-336px~\cite{llava15} & \textit{12.8B}+558K & 665K & 78.5 & 62.0 & 66.8 & 58.2 & 85.9 & 64.3 & 66.1 & 28.1 & 18.2 & 54.8 & 32.4 & 58.3  \\
        \model & 512M & 665K &  77.8 & 61.6 & 68.0 & 56.4 & 86.6 & 61.3 & 69.8 & 29.3 & 21.5 & 58.7 & 37.1 & 59.1 \\
        \bottomrule
        \end{tabular}
    }

    \caption{
    \textbf{Comparison of \model and LLaVA1.5.} We evaluate the models on VQAv2~\cite{vqav2}, GQA~\cite{hudson2019gqa},
    ScienceQA~\cite{scienceqa}, TextVQA~\cite{textvqa}, POPE~\cite{POPE}, MMBench~\cite{liu2024mmbench}, SEEDBench~\cite{seedbench}, DocVQA~\cite{mathew2021docvqa}, ChartQA~\cite{masry2022chartqa}, AI2D~\cite{AI2D} and MMStar~\cite{mmstar}.
    }
    \label{tab:exp_cmp_llava}
\end{table*}

\section*{Appendix}
In the appendix, we provide additional experimental details and results.

\subsection*{Additional Experimental Details}
\label{sec:supp_exp_details}
\head{Training Configurations.}
In this section, we provide the corresponding setups for our experiment series in the main paper, including the default setting, the data scaling series, the model scaling series, and ablation experiment settings. The detailed configurations are shown in Table~\ref{tab:supp_exp_config}.

\head{Evaluation Configurations.}
In the main paper, we measure the model performance on several benchmarks: VQAv2~\cite{vqav2}, GQA~\cite{hudson2019gqa}, ScienceQA-IMG~\cite{scienceqa}, TextVQA~\cite{textvqa}, POPE~\cite{POPE}, MME~\cite{mme}, MMBench~\cite{liu2024mmbench}, and SEEDBench-IMG~\cite{seedbench}. We normalized the performance to a full score of 100 and averaged the performance across these benchmarks to plot the curves shown in Figure~\ref{fig:teaser}(A) and Figure~\ref{fig:exp_modeling_scaling}. The detailed experimental results are shown in Table~\ref{tab:supp_detailed_number}.

\subsection*{Additional Experimental Results}

\head{A comparison of \model and LLaVA1.5.}
In this section, we conduct an experiment to compare \model with LLaVA1.5~\cite{llava15}. In this experiment, our \model is trained on 512M image-text pairs in Pretraining Stage 1, followed by fine-tuning on the LLaVA-mix-665K dataset. To fairly compare the performance of the two models, we do not use the anyres strategy during SFT. Instead, we adopt the same image processing approach as LLaVA1.5, ensuring that the aspect ratio and number of image tokens are consistent across both models.

The experimental results are presented in Table~\ref{tab:exp_cmp_llava}. Despite our model being trained on only 512M image-text pairs, which is significantly smaller than the CLIP pretraining data used in the LLaVA1.5 model, the results show that our model achieves comparable performance to LLaVA1.5 across various benchmarks. Remarkably, our model even outperforms LLaVA1.5 on specific benchmarks such as DocVQA and ChartVQA.

These findings highlight the strong potential of Single Transformer models in terms of data scaling. Specifically, they suggest that even with a relatively smaller pretraining dataset, Single Transformer models can perform on par with, or even exceed, more extensively trained modular MLLMs like LLaVA1.5 when similar preprocessing and controlled variables are applied.

\begin{figure*}[htp]
    \centering
    \includegraphics[width=0.85\textwidth]{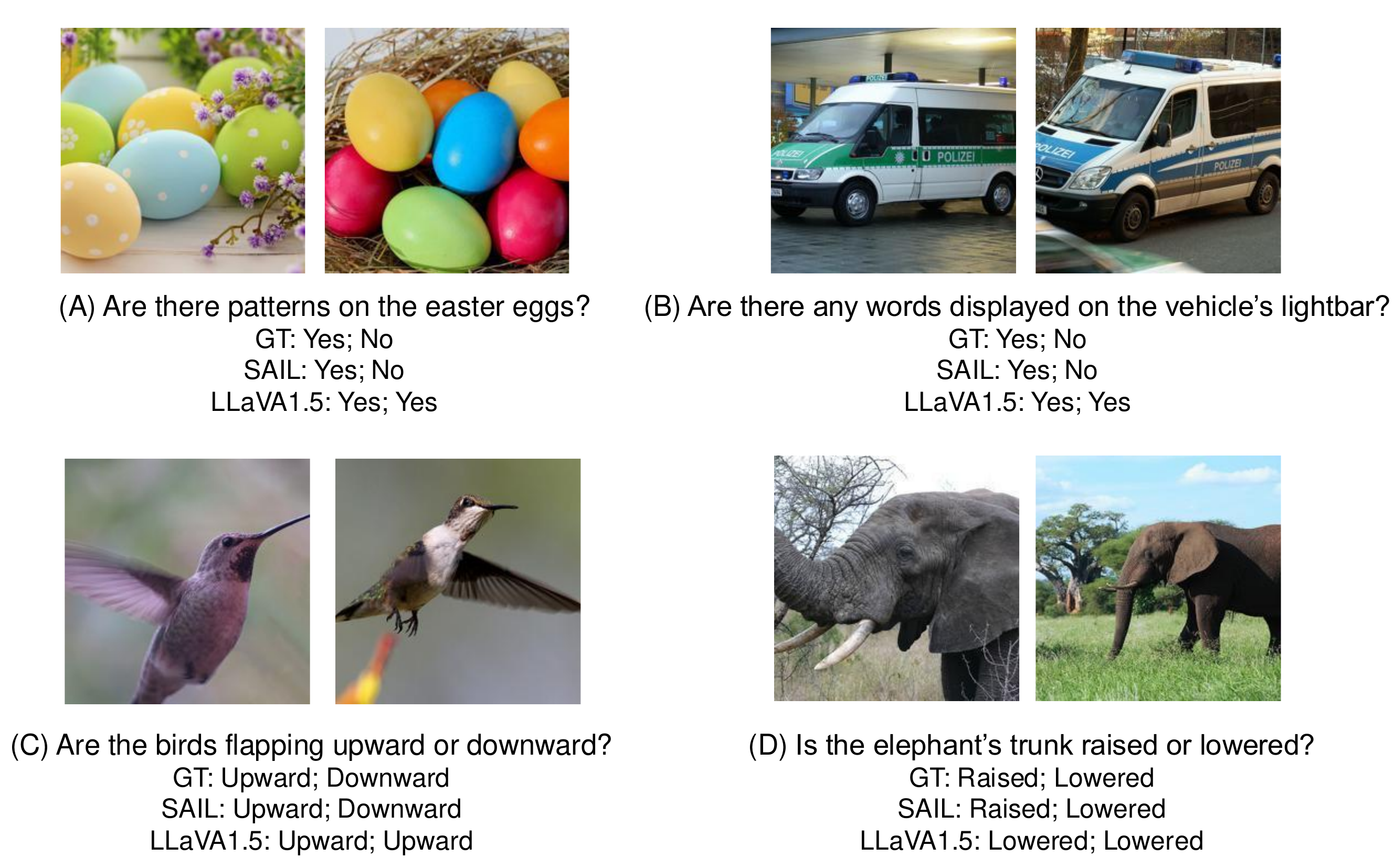}
    \caption{Comparison of \model and LLaVA1.5 on MMVP examples. \model demonstrates better performance in perceiving minor regions and objects, as well as more accurately distinguishing object states.}
    
    \label{fig:supp_mmvp_example}
    \vspace{-1em}
\end{figure*}

\head{Compare \model and LLaVA on MMVP.} We compare \model and LLaVA1.5~\cite{llava15} on MMVP~\cite{mmvp} to dissect the behavior of the two models. The results are shown in Figure~\ref{fig:supp_mmvp_example}. From examples (A) and (B), we observe that \model performs better in perceiving minor regions and objects. Examples (C) and (D) illustrate that \model can more accurately distinguish the states of objects.

\begin{figure*}[htp]
    \centering
    \includegraphics[width=0.6\textwidth]{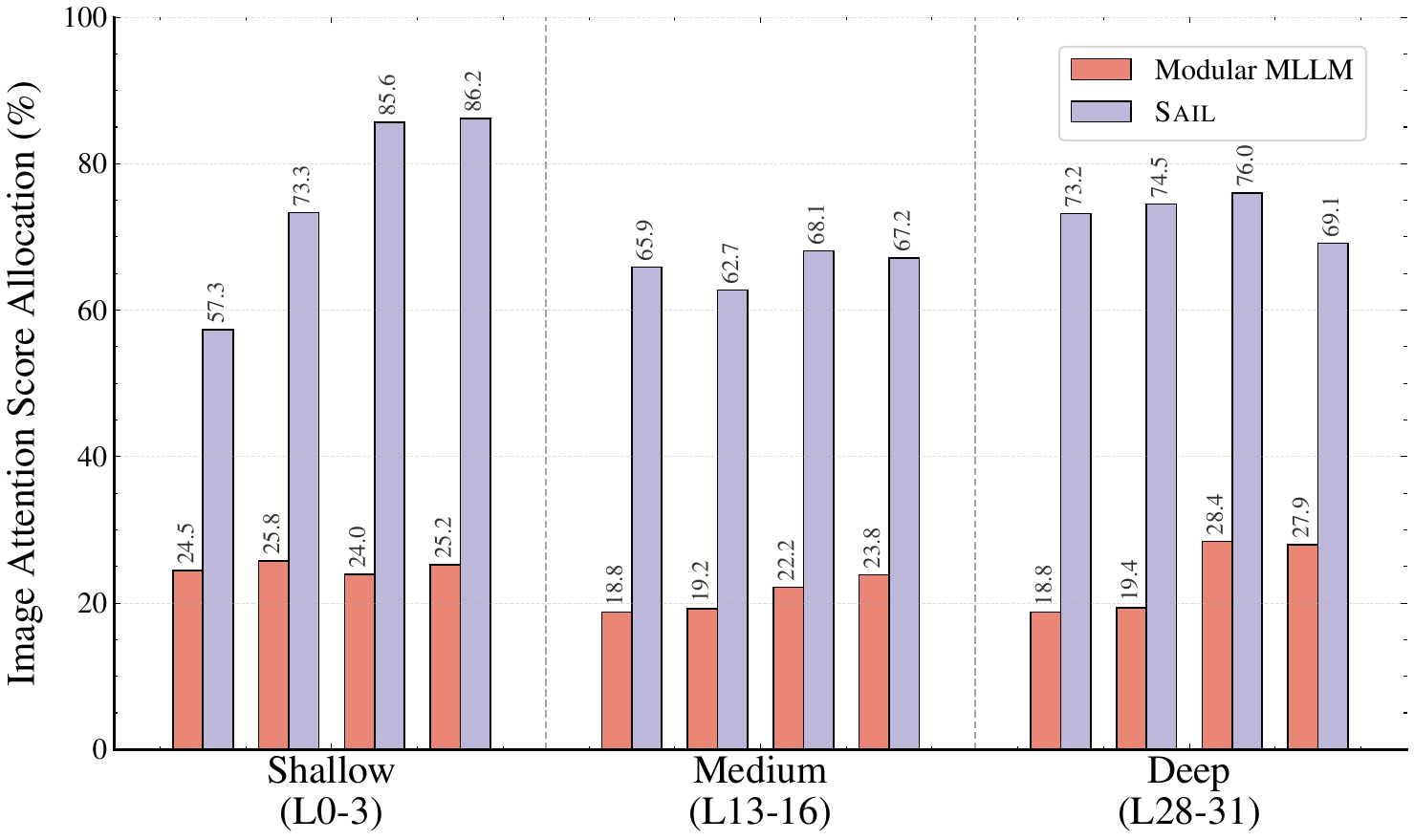}
    \caption{
        \textbf{Image attention score allocation for \model and its modular MLLM counterpart.} 
        We compared the attention score allocation distribution for shallow layers, medium layers, and deep layers between these two models. 
        The Single Transformer-based MLLM model significantly allocates a higher proportion of attention score to image tokens during prediction than the modular MLLM.
    }
    \label{fig:exp_layer_attn_cmp}
\end{figure*}

\head{Additional Experiments on Information Flow Pattern Analysis.}
In the main paper, we analyzed the distribution patterns of image attention scores for different Single Transformer-based MLLMs and modular MLLMs. The results showed that Single Transformer-based MLLMs allocate more attention weights to image tokens. However, this could be due to different models processing varying numbers of image tokens, where more image tokens lead to higher aggregated attention scores.

To analyze this in a more controlled manner, we designed an additional experiment. Using the data scaling setup at 512M, we pretrained \model and its modular MLLM counterpart. After pretraining, we fine-tuned both models using the LLaVA-mix-665K dataset, fixing the resolution size to 224x224 during SFT, instead of using any resolution.

The results, shown in Figure~\ref{fig:exp_layer_attn_cmp}, reveal that \model allocates higher attention scores to image tokens across all transformer layers compared to the modular MLLM, particularly in medium layers (+43.5\% in layer 14) and deep layers (+41.2\% in layer 31).

From this, we can conclude that Single Transformer-based MLLMs tend to allocate a significant portion of attention to previous image tokens during prediction. In contrast, modular MLLMs allocate a smaller portion of their attention directly to image tokens, indicating a less image-centric approach in their prediction mechanism.

\head{Attention Map Visualization.}
In the main paper, we found that Single Transformer-based MLLMs allocate a large portion of attention weights to image tokens during inference, indicating a more vision-centric model. Here, we visualize the attention distribution of \model across different regions of the image when predicting tokens.

\begin{figure*}[htp]
    \centering
    \includegraphics[width=0.7\textwidth]{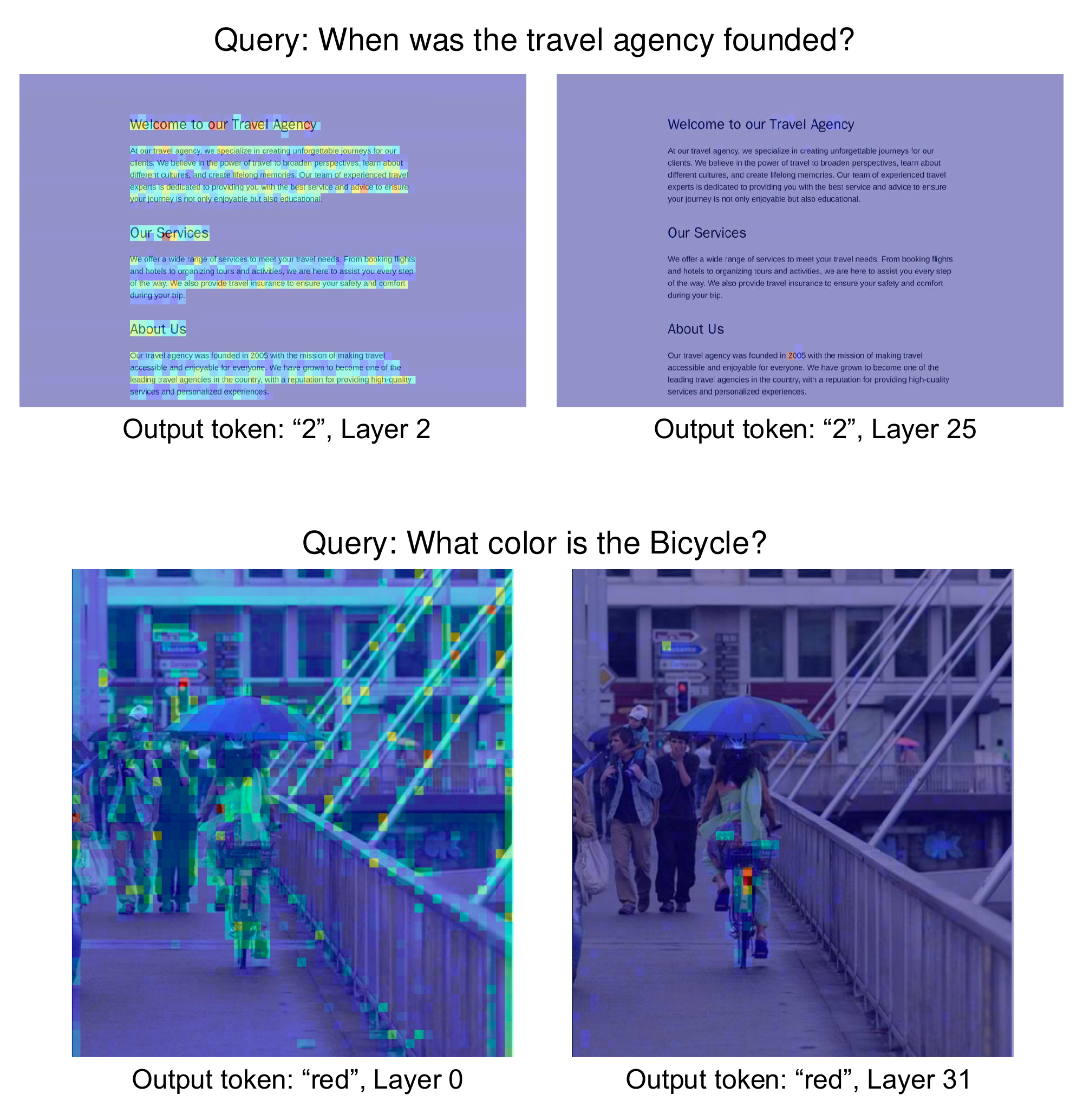}
    \caption{
    \textbf{Visualization of \model's attention distribution across image regions during token prediction.} 
    In early transformer layers, attention primarily focuses on the salient regions of the image. As the model progresses to deeper layers, attention shifts to areas more relevant to the predicted tokens.
    }
    \label{fig:supp_attn_vis}
    \vspace{-1em}
\end{figure*}

The results in Figure~\ref{fig:supp_attn_vis} illustrate the attention maps for specific tokens to the image portion when \model generates predictions for multimodal queries. The visualizations show that in the early transformer layers, the predicted tokens primarily focus on the salient regions of the image. As the model progresses to deeper layers, the attention shifts to areas more relevant to the predicted tokens. This behavior demonstrates that \model has the potential to function as a grounding model, effectively correlating text tokens with their corresponding image regions.

In other words, during inference, the model incrementally concentrates attention weights on relevant regions, aiding in decision-making. This progressive focusing of attention signifies the model's capability to ground text tokens in the corresponding visual context, enhancing its performance in vision-language tasks.

\head{Visual Understanding Demonstration.} 
We investigate several vision perception and reasoning capabilities of our \model. These include its ability to understand rich OCR information (Table~\ref{tab:supp_vis_ocr}), interpret real-world scenes (Table~\ref{tab:supp_vis_realworld}), comprehend scientific charts (Table~\ref{tab:supp_vis_scientific}), and analyze poster contents (Table~\ref{tab:supp_vis_poster}).

\clearpage
\begin{table*}
\begin{minipage}{0.99\textwidth}
\begin{AIbox}{Example 1: Understanding OCR Information in Webpage.}
\centering
\scalebox{0.80}{
\begin{tabular}{l p{18cm}}
&  \includegraphics[width=16cm]{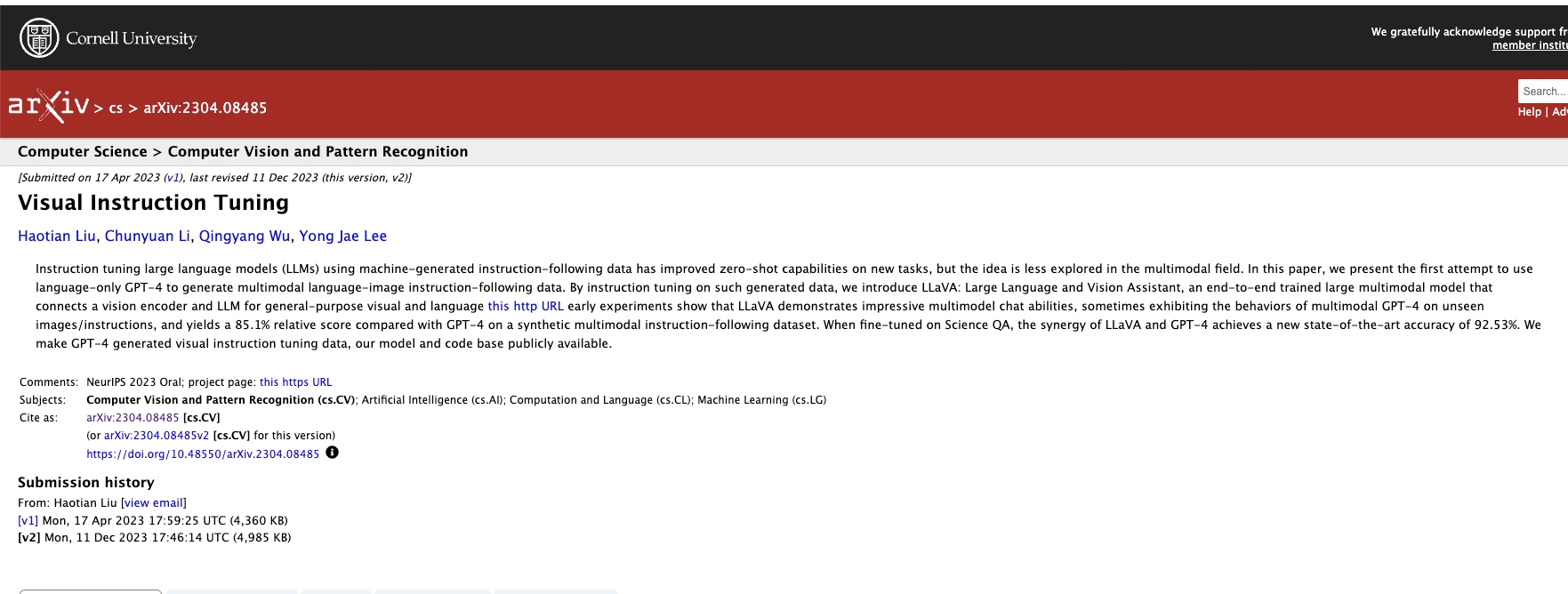} \\
\footnotesize
\textbf{User} & Please output all the text information. \\
\midrule
\footnotesize
\textbf{\model} & Cornell University arXiv cs arXiv:2304.08485
Computer Science Computer Vision and Pattern Recognition

Visual Instruction Tuning HaoTian Liu, Chunyuan Li, Qingyang Wu, Yong Jae Lee Instruction tuning large language models (LLMs) using machine-generated instruction-following data has improved zero-shot capabilities on new tasks, but the idea is less explored in the multimodal field. In this paper, we present the first attempt to use language-only GPT-4 to generate multimodal language-image instruction-following data. By instruction tuning on such generated data, we introduce LLM for general-purpose visual and language visual and language this http URL URL Early Experiments: LLMs demonstrate impressive multimodel chat abilities, sometimes exhibiting the behaviors of multimodal GPT-4 on a synthetic multimodal instruction-following dataset. When fine-tuned on Science QA, the synergy of LLAVA and GPT-4 achieves a new state-of-the-art accuracy of 92.53$\%$. We make GPT-4 generated visual instruction tuning data, our model and code base publicly available.
Comments: NeurIPS 2023 Oral; project page: this https URL
Subjects: Computer Vision and Pattern Recognition (cs.CV); Artificial Intelligence (cs.AI); Computation and Language (cs.CL); Machine Learning (cs.LG)
Cite as: arXiv:2304.08485 [cs.CV] (or arXiv:2304.08485v2 [cs.CV] for this version)
https://doi.org/10.48550/arXiv.2304.08485
Submission history
From: Haotian Liu [view email]
[v1] Mon, 17 Apr 2023 17:59:25 UTC (4,360 KB)
[v2] Mon, 11 Dec 2023 17:46:14 UTC (4,985 KB)
\\
\end{tabular}
}
\end{AIbox}
\captionof{table}{\model demonstrates a strong capability to extract information in OCR-rich scenarios.}
\vspace{-1mm}
\label{tab:supp_vis_ocr}
\end{minipage}
\end{table*}

\begin{table*}
\begin{minipage}{0.99\textwidth}
\begin{AIbox}{Example 2: Understanding Real-World Scene.}
\centering
\scalebox{0.80}{
\begin{tabular}{l p{18cm}}
&  \includegraphics[width=16cm]{figure/realworld.jpg} \\
\footnotesize
\textbf{User} & Please describe the image in detail. \\
\midrule
\footnotesize
\textbf{\model} & There are several notable elements in the image:

1. The highway is multi-lane, with multiple lanes visible.
2. The traffic is dense, with vehicles in various lanes.
3. The weather seems to be overcast, as indicated by the grayish tones of the image.
4. There are several traffic signs and signs visible, including a green highway sign indicating ``1/2 mile'' to the right, and a green sign with the number ``16'' and the word ``MILE'' on the left side of the image.
5. The background includes a hill or mountain, with trees and greenery visible.
6. There are also several power lines and electrical pylons visible along the sides of the highway. 
The image's focus is on the traffic and the surrounding environment. The vehicles are primarily in the center and right lanes of the highway, with some vehicles in the left lanes as well. The highway is flanked by lush greenery, including trees and grassy areas, suggesting that this is a rural or possibly a suburban area.
The traffic appears to be moving at a moderate pace. The weather conditions suggest that this is a daytime scene, possibly during the morning or late afternoon.
The image is taken from a perspective that suggests the viewer is in a vehicle on the highway, observing the traffic ahead.
In summary, the image captures a typical highway scene with moderate traffic, surrounded by a mix of natural and built environments, under an overcast sky. The road is in good condition, and the vehicles are traveling smoothly in both directions.
\\
\end{tabular}
}
\end{AIbox}
\captionof{table}{\model demonstrates a strong capability about understanding the real-world scene.}
\vspace{-1mm}
\label{tab:supp_vis_realworld}
\end{minipage}
\end{table*}

\begin{table*}
\begin{minipage}{0.99\textwidth}
\begin{AIbox}{Example 3: Understanding the Scientific Charts.}
\centering
\scalebox{0.80}{
\begin{tabular}{l p{18cm}}
&  \includegraphics[width=16cm]{figure/ai2d_img.jpg} \\
\footnotesize
\textbf{User} & Please describe the image in detail. \\
\midrule
\footnotesize
\textbf{\model} & 
This image is a collage of various objects, foods, and animals, all of which are predominantly blue or have blue elements. The image is divided into sections labeled with letters from A to O. Here is a detailed description of each section: 1. A - Pumpkin: A blue pumpkin is depicted, which is a common symbol associated with autumn and Halloween. 2. B - Fish: A blue fish is shown, which could be a representation of a common aquatic animal. 3. C - M\&M: A blue M\&M candy is present.4. D - Milk and Soap: A glass of milk, a bar of soap, and a blue sponge.5. E - Bread and Jelly Beans: Blue wheat, a blue loaf of bread, and blue jelly beans.6. F - Castor Oil: A bottle of castor oil.7. G - Cheerios and Snacks: A box of Cheerios, a glass of a dark blue drink, a blue popsicle, a blue donut, and a blue muffin.
8. H - Vegetables: A collection of vegetables including a blue cucumber, green leafy vegetables, and a bunch of blue carrots.9. I - Apple and Stones: A green apple, a blue candy, and blue stones.10. J - Syrup and Candy: Sugar cubes, a bottle of blue syrup, and a blue candy bar.11. K - Coconut and Aloe: A blue coconut, a bottle of blue liquid, and a blue aloe vera plant.12. L - Pineapple and Juice: A blue pineapple, a blue banana, and a glass of blue juice.13. M - Candy Bar: A blue candy bar.14. N - Blueberries: Blueberries.15. O - Oval Object: A blue oval-shaped object.
\\
\end{tabular}
}
\end{AIbox}
\captionof{table}{\model demonstrates a strong capability about understanding the scientific charts.}
\vspace{-1mm}
\label{tab:supp_vis_scientific}
\end{minipage}
\end{table*}

\begin{table*}
\begin{minipage}{0.99\textwidth}
\begin{AIbox}{Example 4: Understanding Poster Content.}
\centering
\scalebox{0.80}{
\begin{tabular}{c p{18cm}}
&  \includegraphics[width=11cm]{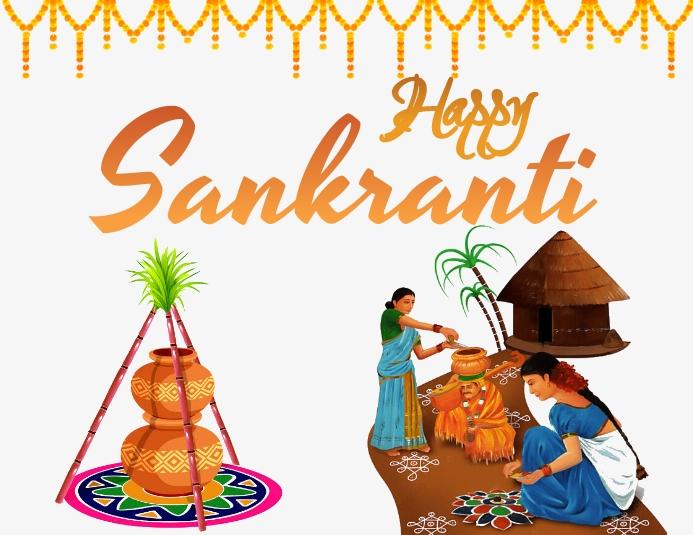} \\
\footnotesize
\textbf{User} & Please describe the image in detail. \\
\midrule
\footnotesize
\textbf{\model} & 
At the top of the image, there is a decorative border with a pattern of hanging golden ornaments, which adds a festive and celebratory feel to the overall design. The background is plain white, which helps the colorful elements stand out.

In the center of the image, the text ``Happy Sankranti'' is prominently displayed in a large, elegant, and stylized font. The text is in a warm orange color, which is traditionally associated with the festival.

Below the text, there is a depiction of a woman dressed in traditional Indian attire, likely a saree, in shades of blue and green. She is engaged in the act of preparing a traditional Sankranti sweet, which is a common practice during the festival. The sweet is shaped like a pyramid and is decorated with intricate patterns, reflecting the artistic and cultural heritage of the region.

To the left of the woman, there is a three-tiered pot, known as a ``patala'', which is a traditional Indian cooking vessel. The pot is adorned with a colorful design and is placed on a colorful circular mat, which is also a traditional element in Indian households.

On the right side of the woman, there is a small hut with a thatched roof, which is a common architectural style in rural Indian villages. The hut is depicted in a simple and rustic manner, emphasizing the rural and traditional aspect of the scene.

Overall, the image captures the essence of the Sankranti festival, highlighting the cultural and religious aspects of the celebration. The use of traditional clothing, the preparation of traditional sweets, and the depiction of a rural village scene all contribute to a sense of cultural authenticity and celebration.

\\
\end{tabular}
}
\end{AIbox}
\captionof{table}{\model demonstrates a strong capability about understanding the poster content.}
\vspace{-1mm}
\label{tab:supp_vis_poster}
\end{minipage}
\end{table*}
\clearpage
{
    \small
    \bibliographystyle{ieeenat_fullname}
    \bibliography{main}

\begin{thebibliography}{91}
\providecommand{\natexlab}[1]{#1}
\providecommand{\url}[1]{\texttt{#1}}
\expandafter\ifx\csname urlstyle\endcsname\relax
  \providecommand{\doi}[1]{doi: #1}\else
  \providecommand{\doi}{doi: \begingroup \urlstyle{rm}\Url}\fi

\bibitem[Alayrac et~al.(2022)Alayrac, Donahue, Luc, Miech, Barr, Hasson, Lenc, Mensch, Millican, Reynolds, et~al.]{alayrac2022flamingo}
Jean-Baptiste Alayrac, Jeff Donahue, Pauline Luc, Antoine Miech, Iain Barr, Yana Hasson, Karel Lenc, Arthur Mensch, Katherine Millican, Malcolm Reynolds, et~al.
\newblock Flamingo: a visual language model for few-shot learning.
\newblock \emph{Advances in neural information processing systems}, 35:\penalty0 23716--23736, 2022.

\bibitem[Anthropic(2024)]{claude}
AI Anthropic.
\newblock The claude 3 model family: Opus, sonnet, haiku.
\newblock \emph{Claude-3 Model Card}, 2024.

\bibitem[Bai et~al.(2023)Bai, Bai, Yang, Wang, Tan, Wang, Lin, Zhou, and Zhou]{bai2023qwen-vl}
Jinze Bai, Shuai Bai, Shusheng Yang, Shijie Wang, Sinan Tan, Peng Wang, Junyang Lin, Chang Zhou, and Jingren Zhou.
\newblock Qwen-vl: A frontier large vision-language model with versatile abilities.
\newblock \emph{arXiv preprint arXiv:2308.12966}, 2023.

\bibitem[Bao et~al.(2021)Bao, Dong, Piao, and Wei]{bao2021beit}
Hangbo Bao, Li Dong, Songhao Piao, and Furu Wei.
\newblock Beit: Bert pre-training of image transformers.
\newblock \emph{arXiv preprint arXiv:2106.08254}, 2021.

\bibitem[Bavishi et~al.(2023)Bavishi, Elsen, Hawthorne, Nye, Odena, Somani, and Ta\c{s}\i{}rlar]{fuyu-8b}
Rohan Bavishi, Erich Elsen, Curtis Hawthorne, Maxwell Nye, Augustus Odena, Arushi Somani, and Sa\u{g}nak Ta\c{s}\i{}rlar.
\newblock Introducing our multimodal models, 2023.

\bibitem[Changpinyo et~al.(2021)Changpinyo, Sharma, Ding, and Soricut]{cc12m}
Soravit Changpinyo, Piyush Sharma, Nan Ding, and Radu Soricut.
\newblock Conceptual 12m: Pushing web-scale image-text pre-training to recognize long-tail visual concepts.
\newblock In \emph{Proceedings of the IEEE/CVF conference on computer vision and pattern recognition}, pages 3558--3568, 2021.

\bibitem[Chen et~al.(2023)Chen, Zhu, Shen, Li, Liu, Zhang, Krishnamoorthi, Chandra, Xiong, and Elhoseiny]{chen2023minigptv2}
Jun Chen, Deyao Zhu, Xiaoqian Shen, Xiang Li, Zechun Liu, Pengchuan Zhang, Raghuraman Krishnamoorthi, Vikas Chandra, Yunyang Xiong, and Mohamed Elhoseiny.
\newblock Minigpt-v2: large language model as a unified interface for vision-language multi-task learning.
\newblock \emph{arXiv preprint arXiv:2310.09478}, 2023.

\bibitem[Chen et~al.(2024{\natexlab{a}})Chen, Yu, Shen, Yuille, and Chen]{chen2024vitamin}
Jieneng Chen, Qihang Yu, Xiaohui Shen, Alan Yuille, and Liang-Chieh Chen.
\newblock Vitamin: Designing scalable vision models in the vision-language era.
\newblock In \emph{Proceedings of the IEEE/CVF Conference on Computer Vision and Pattern Recognition}, 2024{\natexlab{a}}.

\bibitem[Chen et~al.(2024{\natexlab{b}})Chen, Li, Dong, Zhang, Zang, Chen, Duan, Wang, Qiao, Lin, et~al.]{mmstar}
Lin Chen, Jinsong Li, Xiaoyi Dong, Pan Zhang, Yuhang Zang, Zehui Chen, Haodong Duan, Jiaqi Wang, Yu Qiao, Dahua Lin, et~al.
\newblock Are we on the right way for evaluating large vision-language models?
\newblock \emph{arXiv preprint arXiv:2403.20330}, 2024{\natexlab{b}}.

\bibitem[Chen et~al.(2024{\natexlab{c}})Chen, Zhao, Liu, Bai, Lin, Zhou, and Chang]{fastv}
Liang Chen, Haozhe Zhao, Tianyu Liu, Shuai Bai, Junyang Lin, Chang Zhou, and Baobao Chang.
\newblock An image is worth 1/2 tokens after layer 2: Plug-and-play inference acceleration for large vision-language models.
\newblock In \emph{European Conference on Computer Vision}, pages 19--35. Springer, 2024{\natexlab{c}}.

\bibitem[Chen et~al.(2024{\natexlab{d}})Chen, Wang, Peng, and Ji]{chen2024SOLO}
Yangyi Chen, Xingyao Wang, Hao Peng, and Heng Ji.
\newblock A single transformer for scalable vision-language modeling.
\newblock \emph{Transactions on Machine Learning Research}, 2024{\natexlab{d}}.

\bibitem[Chen et~al.(2022)Chen, Duan, Wang, He, Lu, Dai, and Qiao]{chen2022vitadapter}
Zhe Chen, Yuchen Duan, Wenhai Wang, Junjun He, Tong Lu, Jifeng Dai, and Yu Qiao.
\newblock Vision transformer adapter for dense predictions.
\newblock \emph{arXiv preprint arXiv:2205.08534}, 2022.

\bibitem[Chen et~al.(2024{\natexlab{e}})Chen, Wang, Tian, Ye, Gao, Cui, Tong, Hu, Luo, Ma, et~al.]{chen2024far}
Zhe Chen, Weiyun Wang, Hao Tian, Shenglong Ye, Zhangwei Gao, Erfei Cui, Wenwen Tong, Kongzhi Hu, Jiapeng Luo, Zheng Ma, et~al.
\newblock How far are we to gpt-4v? closing the gap to commercial multimodal models with open-source suites.
\newblock \emph{Science China Information Sciences}, 67\penalty0 (12):\penalty0 220101, 2024{\natexlab{e}}.

\bibitem[Chen et~al.(2024{\natexlab{f}})Chen, Wu, Wang, Su, Chen, Xing, Zhong, Zhang, Zhu, Lu, et~al.]{chen2024internvl}
Zhe Chen, Jiannan Wu, Wenhai Wang, Weijie Su, Guo Chen, Sen Xing, Muyan Zhong, Qinglong Zhang, Xizhou Zhu, Lewei Lu, et~al.
\newblock Internvl: Scaling up vision foundation models and aligning for generic visual-linguistic tasks.
\newblock In \emph{Proceedings of the IEEE/CVF Conference on Computer Vision and Pattern Recognition}, pages 24185--24198, 2024{\natexlab{f}}.

\bibitem[Cherti et~al.(2023)Cherti, Beaumont, Wightman, Wortsman, Ilharco, Gordon, Schuhmann, Schmidt, and Jitsev]{openclip}
Mehdi Cherti, Romain Beaumont, Ross Wightman, Mitchell Wortsman, Gabriel Ilharco, Cade Gordon, Christoph Schuhmann, Ludwig Schmidt, and Jenia Jitsev.
\newblock Reproducible scaling laws for contrastive language-image learning.
\newblock In \emph{Proceedings of the IEEE/CVF Conference on Computer Vision and Pattern Recognition}, pages 2818--2829, 2023.

\bibitem[Chiang et~al.(2023)Chiang, Li, Lin, Sheng, Wu, Zhang, Zheng, Zhuang, Zhuang, Gonzalez, Stoica, and Xing]{vicuna2023}
Wei-Lin Chiang, Zhuohan Li, Zi Lin, Ying Sheng, Zhanghao Wu, Hao Zhang, Lianmin Zheng, Siyuan Zhuang, Yonghao Zhuang, Joseph~E. Gonzalez, Ion Stoica, and Eric~P. Xing.
\newblock Vicuna: An open-source chatbot impressing gpt-4 with 90\%* chatgpt quality, 2023.

\bibitem[Contributors(2020)]{mmseg2020}
MMSegmentation Contributors.
\newblock {MMSegmentation}: Openmmlab semantic segmentation toolbox and benchmark.
\newblock \url{https://github.com/open-mmlab/mmsegmentation}, 2020.

\bibitem[Dai et~al.(2023)Dai, Li, Li, Tiong, Zhao, Wang, Li, Fung, and Hoi]{dai2023instructblip}
Wenliang Dai, Junnan Li, Dongxu Li, Anthony Tiong, Junqi Zhao, Weisheng Wang, Boyang Li, Pascale Fung, and Steven Hoi.
\newblock Instruct{BLIP}: Towards general-purpose vision-language models with instruction tuning.
\newblock In \emph{Thirty-seventh Conference on Neural Information Processing Systems}, 2023.

\bibitem[Dehghani et~al.(2023)Dehghani, Djolonga, Mustafa, Padlewski, Heek, Gilmer, Steiner, Caron, Geirhos, Alabdulmohsin, et~al.]{dehghani2023scaling}
Mostafa Dehghani, Josip Djolonga, Basil Mustafa, Piotr Padlewski, Jonathan Heek, Justin Gilmer, Andreas~Peter Steiner, Mathilde Caron, Robert Geirhos, Ibrahim Alabdulmohsin, et~al.
\newblock Scaling vision transformers to 22 billion parameters.
\newblock In \emph{International Conference on Machine Learning}, pages 7480--7512. PMLR, 2023.

\bibitem[Deng et~al.(2009)Deng, Dong, Socher, Li, Li, and Fei-Fei]{deng2009imagenet}
Jia Deng, Wei Dong, Richard Socher, Li-Jia Li, Kai Li, and Li Fei-Fei.
\newblock Imagenet: A large-scale hierarchical image database.
\newblock In \emph{2009 IEEE conference on computer vision and pattern recognition}, pages 248--255. Ieee, 2009.

\bibitem[Diao et~al.(2024)Diao, Cui, Li, Wang, Lu, and Wang]{diao2024EVE}
Haiwen Diao, Yufeng Cui, Xiaotong Li, Yueze Wang, Huchuan Lu, and Xinlong Wang.
\newblock Unveiling encoder-free vision-language models.
\newblock In \emph{Advances in Neural Information Processing Systems}, pages 52545--52567. Curran Associates, Inc., 2024.

\bibitem[Diao et~al.(2025)Diao, Li, Cui, Wang, Deng, Pan, Wang, Lu, and Wang]{diao2025EVEv2}
Haiwen Diao, Xiaotong Li, Yufeng Cui, Yueze Wang, Haoge Deng, Ting Pan, Wenxuan Wang, Huchuan Lu, and Xinlong Wang.
\newblock Evev2: Improved baselines for encoder-free vision-language models.
\newblock \emph{arXiv preprint arXiv:2502.06788}, 2025.

\bibitem[Dosovitskiy et~al.(2021)Dosovitskiy, Beyer, Kolesnikov, Weissenborn, Zhai, Unterthiner, Dehghani, Minderer, Heigold, Gelly, Uszkoreit, and Houlsby]{dosovitskiy2021vit}
Alexey Dosovitskiy, Lucas Beyer, Alexander Kolesnikov, Dirk Weissenborn, Xiaohua Zhai, Thomas Unterthiner, Mostafa Dehghani, Matthias Minderer, Georg Heigold, Sylvain Gelly, Jakob Uszkoreit, and Neil Houlsby.
\newblock An image is worth 16x16 words: Transformers for image recognition at scale.
\newblock In \emph{International Conference on Learning Representations}, 2021.

\bibitem[Duan et~al.(2024)Duan, Yang, Qiao, Fang, Chen, Liu, Dong, Zang, Zhang, Wang, et~al.]{duan2024vlmevalkit}
Haodong Duan, Junming Yang, Yuxuan Qiao, Xinyu Fang, Lin Chen, Yuan Liu, Xiaoyi Dong, Yuhang Zang, Pan Zhang, Jiaqi Wang, et~al.
\newblock Vlmevalkit: An open-source toolkit for evaluating large multi-modality models.
\newblock In \emph{Proceedings of the 32nd ACM International Conference on Multimedia}, pages 11198--11201, 2024.

\bibitem[El-Nouby et~al.(2024)El-Nouby, Klein, Zhai, Bautista, Toshev, Shankar, Susskind, and Joulin]{el2024aim}
Alaaeldin El-Nouby, Michal Klein, Shuangfei Zhai, Miguel~Angel Bautista, Alexander Toshev, Vaishaal Shankar, Joshua~M Susskind, and Armand Joulin.
\newblock Scalable pre-training of large autoregressive image models.
\newblock \emph{arXiv preprint arXiv:2401.08541}, 2024.

\bibitem[Fang et~al.(2023)Fang, Wang, Xie, Sun, Wu, Wang, Huang, Wang, and Cao]{fang2023eva}
Yuxin Fang, Wen Wang, Binhui Xie, Quan Sun, Ledell Wu, Xinggang Wang, Tiejun Huang, Xinlong Wang, and Yue Cao.
\newblock Eva: Exploring the limits of masked visual representation learning at scale.
\newblock In \emph{Proceedings of the IEEE/CVF conference on computer vision and pattern recognition}, pages 19358--19369, 2023.

\bibitem[Gadre et~al.(2023)Gadre, Ilharco, Fang, Hayase, Smyrnis, Nguyen, Marten, Wortsman, Ghosh, Zhang, et~al.]{datacomp1b}
Samir~Yitzhak Gadre, Gabriel Ilharco, Alex Fang, Jonathan Hayase, Georgios Smyrnis, Thao Nguyen, Ryan Marten, Mitchell Wortsman, Dhruba Ghosh, Jieyu Zhang, et~al.
\newblock Datacomp: In search of the next generation of multimodal datasets.
\newblock \emph{Advances in Neural Information Processing Systems}, 36:\penalty0 27092--27112, 2023.

\bibitem[Goyal et~al.(2017)Goyal, Khot, Summers-Stay, Batra, and Parikh]{vqav2}
Yash Goyal, Tejas Khot, Douglas Summers-Stay, Dhruv Batra, and Devi Parikh.
\newblock Making the v in vqa matter: Elevating the role of image understanding in visual question answering.
\newblock In \emph{Proceedings of the IEEE conference on computer vision and pattern recognition}, pages 6904--6913, 2017.

\bibitem[Gu et~al.(2024)Gu, Zhang, Zhou, Yu, Xing, Wang, Cao, Jia, Zhang, Wang, et~al.]{infinitymm}
Shuhao Gu, Jialing Zhang, Siyuan Zhou, Kevin Yu, Zhaohu Xing, Liangdong Wang, Zhou Cao, Jintao Jia, Zhuoyi Zhang, Yixuan Wang, et~al.
\newblock Infinity-mm: Scaling multimodal performance with large-scale and high-quality instruction data.
\newblock \emph{arXiv preprint arXiv:2410.18558}, 2024.

\bibitem[Guan et~al.(2024)Guan, Liu, Wu, Xian, Li, Liu, Wang, Chen, Huang, Yacoob, et~al.]{hallusionbench}
Tianrui Guan, Fuxiao Liu, Xiyang Wu, Ruiqi Xian, Zongxia Li, Xiaoyu Liu, Xijun Wang, Lichang Chen, Furong Huang, Yaser Yacoob, et~al.
\newblock Hallusionbench: an advanced diagnostic suite for entangled language hallucination and visual illusion in large vision-language models.
\newblock In \emph{Proceedings of the IEEE/CVF Conference on Computer Vision and Pattern Recognition}, pages 14375--14385, 2024.

\bibitem[He et~al.(2020)He, Fan, Wu, Xie, and Girshick]{he2020momentum}
Kaiming He, Haoqi Fan, Yuxin Wu, Saining Xie, and Ross Girshick.
\newblock Momentum contrast for unsupervised visual representation learning.
\newblock In \emph{Proceedings of the IEEE/CVF conference on computer vision and pattern recognition}, pages 9729--9738, 2020.

\bibitem[He et~al.(2022)He, Chen, Xie, Li, Doll{\'a}r, and Girshick]{he2022masked}
Kaiming He, Xinlei Chen, Saining Xie, Yanghao Li, Piotr Doll{\'a}r, and Ross Girshick.
\newblock Masked autoencoders are scalable vision learners.
\newblock In \emph{Proceedings of the IEEE/CVF conference on computer vision and pattern recognition}, pages 16000--16009, 2022.

\bibitem[Hong et~al.(2023)Hong, Wang, Lv, Xu, Yu, Ji, Wang, Wang, Dong, Ding, et~al.]{cogagent}
Wenyi Hong, Weihan Wang, Qingsong Lv, Jiazheng Xu, Wenmeng Yu, Junhui Ji, Yan Wang, Zihan Wang, Yuxiao Dong, Ming Ding, et~al.
\newblock Cogagent: A visual language model for gui agents.
\newblock \emph{arXiv preprint arXiv:2312.08914}, 2023.

\bibitem[Huang et~al.(2023)Huang, Zhang, Ma, Tian, Feng, Zhang, Li, Guo, and Zhang]{huang2023tag2text}
Xinyu Huang, Youcai Zhang, Jinyu Ma, Weiwei Tian, Rui Feng, Yuejie Zhang, Yaqian Li, Yandong Guo, and Lei Zhang.
\newblock Tag2text: Guiding vision-language model via image tagging.
\newblock \emph{arXiv preprint arXiv:2303.05657}, 2023.

\bibitem[Huang et~al.(2024)Huang, Ye, Kang, Feng, and Fan]{superclass_huang}
Zilong Huang, Qinghao Ye, Bingyi Kang, Jiashi Feng, and Haoqi Fan.
\newblock Classification done right for vision-language pre-training.
\newblock In \emph{NeurIPS}, 2024.

\bibitem[Hudson and Manning(2019)]{hudson2019gqa}
Drew~A Hudson and Christopher~D Manning.
\newblock Gqa: A new dataset for real-world visual reasoning and compositional question answering.
\newblock In \emph{Proceedings of the IEEE/CVF conference on computer vision and pattern recognition}, pages 6700--6709, 2019.

\bibitem[Jia et~al.(2021)Jia, Yang, Xia, Chen, Parekh, Pham, Le, Sung, Li, and Duerig]{jia2021scaling}
Chao Jia, Yinfei Yang, Ye Xia, Yi-Ting Chen, Zarana Parekh, Hieu Pham, Quoc Le, Yun-Hsuan Sung, Zhen Li, and Tom Duerig.
\newblock Scaling up visual and vision-language representation learning with noisy text supervision.
\newblock In \emph{International conference on machine learning}, pages 4904--4916. PMLR, 2021.

\bibitem[Jiang et~al.(2023)Jiang, Sablayrolles, Mensch, Bamford, Chaplot, Casas, Bressand, Lengyel, Lample, Saulnier, et~al.]{jiang2023mistral}
Albert~Q Jiang, Alexandre Sablayrolles, Arthur Mensch, Chris Bamford, Devendra~Singh Chaplot, Diego de~las Casas, Florian Bressand, Gianna Lengyel, Guillaume Lample, Lucile Saulnier, et~al.
\newblock Mistral 7b.
\newblock \emph{arXiv preprint arXiv:2310.06825}, 2023.

\bibitem[Kembhavi et~al.(2016)Kembhavi, Salvato, Kolve, Seo, Hajishirzi, and Farhadi]{AI2D}
Aniruddha Kembhavi, Mike Salvato, Eric Kolve, Minjoon Seo, Hannaneh Hajishirzi, and Ali Farhadi.
\newblock A diagram is worth a dozen images, 2016.

\bibitem[Li et~al.(2023{\natexlab{a}})Li, Wang, Wang, Ge, Ge, and Shan]{seedbench}
Bohao Li, Rui Wang, Guangzhi Wang, Yuying Ge, Yixiao Ge, and Ying Shan.
\newblock Seed-bench: Benchmarking multimodal llms with generative comprehension.
\newblock \emph{arXiv preprint arXiv:2307.16125}, 2023{\natexlab{a}}.

\bibitem[Li et~al.(2024{\natexlab{a}})Li, Zhang, Guo, Zhang, Li, Zhang, Zhang, Li, Liu, and Li]{li2024llavaonevision}
Bo Li, Yuanhan Zhang, Dong Guo, Renrui Zhang, Feng Li, Hao Zhang, Kaichen Zhang, Yanwei Li, Ziwei Liu, and Chunyuan Li.
\newblock Llava-onevision: Easy visual task transfer.
\newblock \emph{arXiv preprint arXiv:2408.03326}, 2024{\natexlab{a}}.

\bibitem[Li et~al.(2023{\natexlab{b}})Li, Wang, and Xie]{li2023clipa}
Xianhang Li, Zeyu Wang, and Cihang Xie.
\newblock An inverse scaling law for clip training.
\newblock In \emph{NeurIPS}, 2023{\natexlab{b}}.

\bibitem[Li et~al.(2024{\natexlab{b}})Li, Tu, Hui, Wang, Zhao, Xiao, Ren, Mei, Liu, Zheng, Zhou, and Xie]{recap_datacomp1b}
Xianhang Li, Haoqin Tu, Mude Hui, Zeyu Wang, Bingchen Zhao, Junfei Xiao, Sucheng Ren, Jieru Mei, Qing Liu, Huangjie Zheng, Yuyin Zhou, and Cihang Xie.
\newblock What if we recaption billions of web images with llama-3?
\newblock \emph{arXiv preprint arXiv:2406.08478}, 2024{\natexlab{b}}.

\bibitem[Li et~al.(2023{\natexlab{c}})Li, Du, Zhou, Wang, Zhao, and Wen]{POPE}
Yifan Li, Yifan Du, Kun Zhou, Jinpeng Wang, Wayne~Xin Zhao, and Ji-Rong Wen.
\newblock Evaluating object hallucination in large vision-language models.
\newblock \emph{arXiv preprint arXiv:2305.10355}, 2023{\natexlab{c}}.

\bibitem[Li et~al.(2023{\natexlab{d}})Li, Fan, Hu, Feichtenhofer, and He]{li2023scaling}
Yanghao Li, Haoqi Fan, Ronghang Hu, Christoph Feichtenhofer, and Kaiming He.
\newblock Scaling language-image pre-training via masking.
\newblock In \emph{Proceedings of the IEEE/CVF conference on computer vision and pattern recognition}, pages 23390--23400, 2023{\natexlab{d}}.

\bibitem[Liang et~al.(2024)Liang, Xu, Hong, Shang, Wang, Fu, and Liu]{mme}
Zijing Liang, Yanjie Xu, Yifan Hong, Penghui Shang, Qi Wang, Qiang Fu, and Ke Liu.
\newblock A survey of multimodel large language models.
\newblock In \emph{Proceedings of the 3rd International Conference on Computer, Artificial Intelligence and Control Engineering}, pages 405--409, 2024.

\bibitem[Liu et~al.(2024{\natexlab{a}})Liu, Li, Li, and Lee]{llava15}
Haotian Liu, Chunyuan Li, Yuheng Li, and Yong~Jae Lee.
\newblock Improved baselines with visual instruction tuning.
\newblock In \emph{Proceedings of the IEEE/CVF Conference on Computer Vision and Pattern Recognition}, pages 26296--26306, 2024{\natexlab{a}}.

\bibitem[Liu et~al.(2024{\natexlab{b}})Liu, Li, Li, Li, Zhang, Shen, and Lee]{llava1.6}
Haotian Liu, Chunyuan Li, Yuheng Li, Bo Li, Yuanhan Zhang, Sheng Shen, and Yong~Jae Lee.
\newblock Llava-next: Improved reasoning, ocr, and world knowledge, 2024{\natexlab{b}}.

\bibitem[Liu et~al.(2024{\natexlab{c}})Liu, Li, Wu, and Lee]{llava}
Haotian Liu, Chunyuan Li, Qingyang Wu, and Yong~Jae Lee.
\newblock Visual instruction tuning.
\newblock \emph{Advances in neural information processing systems}, 36, 2024{\natexlab{c}}.

\bibitem[Liu et~al.(2024{\natexlab{d}})Liu, Duan, Zhang, Li, Zhang, Zhao, Yuan, Wang, He, Liu, et~al.]{liu2024mmbench}
Yuan Liu, Haodong Duan, Yuanhan Zhang, Bo Li, Songyang Zhang, Wangbo Zhao, Yike Yuan, Jiaqi Wang, Conghui He, Ziwei Liu, et~al.
\newblock Mmbench: Is your multi-modal model an all-around player?
\newblock In \emph{European conference on computer vision}, pages 216--233. Springer, 2024{\natexlab{d}}.

\bibitem[Liu et~al.(2024{\natexlab{e}})Liu, Li, Huang, Yang, Yu, Li, Yin, Liu, Jin, and Bai]{liu2024ocrbench}
Yuliang Liu, Zhang Li, Mingxin Huang, Biao Yang, Wenwen Yu, Chunyuan Li, Xu-Cheng Yin, Cheng-Lin Liu, Lianwen Jin, and Xiang Bai.
\newblock Ocrbench: on the hidden mystery of ocr in large multimodal models.
\newblock \emph{Science China Information Sciences}, 67\penalty0 (12):\penalty0 220102, 2024{\natexlab{e}}.

\bibitem[Lu et~al.(2022)Lu, Mishra, Xia, Qiu, Chang, Zhu, Tafjord, Clark, and Kalyan]{scienceqa}
Pan Lu, Swaroop Mishra, Tanglin Xia, Liang Qiu, Kai-Wei Chang, Song-Chun Zhu, Oyvind Tafjord, Peter Clark, and Ashwin Kalyan.
\newblock Learn to explain: Multimodal reasoning via thought chains for science question answering.
\newblock \emph{Advances in Neural Information Processing Systems}, 35:\penalty0 2507--2521, 2022.

\bibitem[Lu et~al.(2023)Lu, Bansal, Xia, Liu, Li, Hajishirzi, Cheng, Chang, Galley, and Gao]{lu2023mathvista}
Pan Lu, Hritik Bansal, Tony Xia, Jiacheng Liu, Chunyuan Li, Hannaneh Hajishirzi, Hao Cheng, Kai-Wei Chang, Michel Galley, and Jianfeng Gao.
\newblock Mathvista: Evaluating mathematical reasoning of foundation models in visual contexts.
\newblock \emph{arXiv preprint arXiv:2310.02255}, 2023.

\bibitem[Luo et~al.(2024)Luo, Yang, Dou, Wang, Liu, Dai, Qiao, and Zhu]{luo2024mono-internvl}
Gen Luo, Xue Yang, Wenhan Dou, Zhaokai Wang, Jiawen Liu, Jifeng Dai, Yu Qiao, and Xizhou Zhu.
\newblock Mono-internvl: Pushing the boundaries of monolithic multimodal large language models with endogenous visual pre-training.
\newblock \emph{arXiv preprint arXiv:2410.08202}, 2024.

\bibitem[Masry et~al.(2022)Masry, Long, Tan, Joty, and Hoque]{masry2022chartqa}
Ahmed Masry, Do~Xuan Long, Jia~Qing Tan, Shafiq Joty, and Enamul Hoque.
\newblock Chartqa: A benchmark for question answering about charts with visual and logical reasoning.
\newblock \emph{arXiv preprint arXiv:2203.10244}, 2022.

\bibitem[Mathew et~al.(2021)Mathew, Karatzas, and Jawahar]{mathew2021docvqa}
Minesh Mathew, Dimosthenis Karatzas, and CV Jawahar.
\newblock Docvqa: A dataset for vqa on document images.
\newblock In \emph{Proceedings of the IEEE/CVF winter conference on applications of computer vision}, pages 2200--2209, 2021.

\bibitem[Mehta et~al.(2024)Mehta, Horton, Faghri, Sekhavat, Najibi, Farajtabar, Tuzel, and Rastegari]{mehta2024catlip}
Sachin Mehta, Maxwell Horton, Fartash Faghri, Mohammad~Hossein Sekhavat, Mahyar Najibi, Mehrdad Farajtabar, Oncel Tuzel, and Mohammad Rastegari.
\newblock Catlip: Clip-level visual recognition accuracy with 2.7 x faster pre-training on web-scale image-text data.
\newblock \emph{arXiv preprint arXiv:2404.15653}, 2024.

\bibitem[{Meta}(2024)]{llama3}
{Meta}.
\newblock Introducing meta llama 3: The most capable openly available llm to date, 2024.
\newblock Accessed: 2024-04-18.

\bibitem[OpenAI(2021)]{chatgpt}
OpenAI.
\newblock Introducing chatgpt.
\newblock OpenAI Blog, 2021.

\bibitem[OpenAI(2023{\natexlab{a}})]{gpt4}
OpenAI.
\newblock Gpt-4 technical report, 2023{\natexlab{a}}.

\bibitem[OpenAI(2023{\natexlab{b}})]{gpt4v}
OpenAI.
\newblock Gpt-4v(ision) system card, 2023{\natexlab{b}}.

\bibitem[Radford et~al.(2021)Radford, Kim, Hallacy, Ramesh, Goh, Agarwal, Sastry, Askell, Mishkin, Clark, et~al.]{openai_clip}
Alec Radford, Jong~Wook Kim, Chris Hallacy, Aditya Ramesh, Gabriel Goh, Sandhini Agarwal, Girish Sastry, Amanda Askell, Pamela Mishkin, Jack Clark, et~al.
\newblock Learning transferable visual models from natural language supervision.
\newblock In \emph{International conference on machine learning}, pages 8748--8763. PmLR, 2021.

\bibitem[Schuhmann et~al.(2022)Schuhmann, Beaumont, Vencu, Gordon, Wightman, Cherti, Coombes, Katta, Mullis, Wortsman, et~al.]{schuhmann2022laion5b}
Christoph Schuhmann, Romain Beaumont, Richard Vencu, Cade Gordon, Ross Wightman, Mehdi Cherti, Theo Coombes, Aarush Katta, Clayton Mullis, Mitchell Wortsman, et~al.
\newblock {LAION-5B: An open large-scale dataset for training next generation image-text models}.
\newblock In \emph{NeurIPS}, 2022.

\bibitem[Shoeybi et~al.(2019)Shoeybi, Patwary, Puri, LeGresley, Casper, and Catanzaro]{shoeybi2019megatron}
Mohammad Shoeybi, Mostofa Patwary, Raul Puri, Patrick LeGresley, Jared Casper, and Bryan Catanzaro.
\newblock Megatron-lm: Training multi-billion parameter language models using model parallelism.
\newblock \emph{arXiv preprint arXiv:1909.08053}, 2019.

\bibitem[Singh et~al.(2019)Singh, Natarajan, Shah, Jiang, Chen, Batra, Parikh, and Rohrbach]{textvqa}
Amanpreet Singh, Vivek Natarajan, Meet Shah, Yu Jiang, Xinlei Chen, Dhruv Batra, Devi Parikh, and Marcus Rohrbach.
\newblock Towards vqa models that can read.
\newblock In \emph{Proceedings of the IEEE/CVF conference on computer vision and pattern recognition}, pages 8317--8326, 2019.

\bibitem[Soboleva et~al.(2023)Soboleva, Al-Khateeb, Myers, Steeves, Hestness, and Dey]{cerebras2023slimpajama}
Daria Soboleva, Faisal Al-Khateeb, Robert Myers, Jacob~R Steeves, Joel Hestness, and Nolan Dey.
\newblock {SlimPajama: A 627B token cleaned and deduplicated version of RedPajama}.
\newblock \url{https://cerebras.ai/blog/slimpajama-a-627b-token-cleaned-and-deduplicated-version-of-redpajama}, 2023.

\bibitem[Team(2024)]{chameleon}
Chameleon Team.
\newblock Chameleon: Mixed-modal early-fusion foundation models.
\newblock \emph{arXiv preprint arXiv:2405.09818}, 2024.

\bibitem[Team et~al.(2023)Team, Anil, Borgeaud, Wu, Alayrac, Yu, Soricut, Schalkwyk, Dai, Hauth, et~al.]{gemini}
Gemini Team, Rohan Anil, Sebastian Borgeaud, Yonghui Wu, Jean-Baptiste Alayrac, Jiahui Yu, Radu Soricut, Johan Schalkwyk, Andrew~M Dai, Anja Hauth, et~al.
\newblock Gemini: a family of highly capable multimodal models.
\newblock \emph{arXiv preprint arXiv:2312.11805}, 2023.

\bibitem[Tong et~al.(2024{\natexlab{a}})Tong, Brown, Wu, Woo, IYER, Akula, Yang, Yang, Middepogu, Wang, et~al.]{tong2024cambrian}
Peter Tong, Ellis Brown, Penghao Wu, Sanghyun Woo, Adithya Jairam~Vedagiri IYER, Sai~Charitha Akula, Shusheng Yang, Jihan Yang, Manoj Middepogu, Ziteng Wang, et~al.
\newblock Cambrian-1: A fully open, vision-centric exploration of multimodal llms.
\newblock \emph{Advances in Neural Information Processing Systems}, 37:\penalty0 87310--87356, 2024{\natexlab{a}}.

\bibitem[Tong et~al.(2024{\natexlab{b}})Tong, Liu, Zhai, Ma, LeCun, and Xie]{mmvp}
Shengbang Tong, Zhuang Liu, Yuexiang Zhai, Yi Ma, Yann LeCun, and Saining Xie.
\newblock Eyes wide shut? exploring the visual shortcomings of multimodal llms.
\newblock In \emph{Proceedings of the IEEE/CVF Conference on Computer Vision and Pattern Recognition}, pages 9568--9578, 2024{\natexlab{b}}.

\bibitem[Tschannen et~al.(2023{\natexlab{a}})Tschannen, Kumar, Steiner, Zhai, Houlsby, and Beyer]{CapPa}
Michael Tschannen, Manoj Kumar, Andreas Steiner, Xiaohua Zhai, Neil Houlsby, and Lucas Beyer.
\newblock Image captioners are scalable vision learners too.
\newblock \emph{Advances in Neural Information Processing Systems}, 36:\penalty0 46830--46855, 2023{\natexlab{a}}.

\bibitem[Tschannen et~al.(2023{\natexlab{b}})Tschannen, Kumar, Steiner, Zhai, Houlsby, and Beyer]{tschannen2023cappa}
Michael Tschannen, Manoj Kumar, Andreas Steiner, Xiaohua Zhai, Neil Houlsby, and Lucas Beyer.
\newblock Image captioners are scalable vision learners too.
\newblock \emph{Advances in Neural Information Processing Systems}, 36:\penalty0 46830--46855, 2023{\natexlab{b}}.

\bibitem[Wang et~al.(2023{\natexlab{a}})Wang, Fan, Wang, Song, Wang, Zhang, and Zhang]{wang2023bootstrap}
Haochen Wang, Junsong Fan, Yuxi Wang, Kaiyou Song, Tiancai Wang, Xiangyu Zhang, and Zhaoxiang Zhang.
\newblock Bootstrap masked visual modeling via hard patches mining.
\newblock \emph{arXiv preprint arXiv:2312.13714}, 2023{\natexlab{a}}.

\bibitem[Wang et~al.(2023{\natexlab{b}})Wang, Song, Fan, Wang, Xie, and Zhang]{wang2023hard}
Haochen Wang, Kaiyou Song, Junsong Fan, Yuxi Wang, Jin Xie, and Zhaoxiang Zhang.
\newblock Hard patches mining for masked image modeling.
\newblock In \emph{Proceedings of the IEEE/CVF Conference on Computer Vision and Pattern Recognition}, pages 10375--10385, 2023{\natexlab{b}}.

\bibitem[Wang et~al.(2025)Wang, Zheng, Zhao, Wang, Zheng, Zhang, and Zhang]{wang2025ross}
Haochen Wang, Anlin Zheng, Yucheng Zhao, Tiancai Wang, Ge Zheng, Xiangyu Zhang, and Zhaoxiang Zhang.
\newblock Reconstructive visual instruction tuning.
\newblock In \emph{International Conference on Learning Representations}, 2025.

\bibitem[Wang et~al.(2024{\natexlab{a}})Wang, Wu, Jiang, Xun, Xiao, Guo, and Xiao]{wang2024world}
Jiacong Wang, Bohong Wu, Haiyong Jiang, Zhou Xun, Xin Xiao, Haoyuan Guo, and Jun Xiao.
\newblock World to code: Multi-modal data generation via self-instructed compositional captioning and filtering.
\newblock In \emph{Proceedings of the 2024 Conference on Empirical Methods in Natural Language Processing}, pages 4608--4623, 2024{\natexlab{a}}.

\bibitem[Wang et~al.(2024{\natexlab{b}})Wang, Bai, Tan, Wang, Fan, Bai, Chen, Liu, Wang, Ge, Fan, Dang, Du, Ren, Men, Liu, Zhou, Zhou, and Lin]{qwen2vl}
Peng Wang, Shuai Bai, Sinan Tan, Shijie Wang, Zhihao Fan, Jinze Bai, Keqin Chen, Xuejing Liu, Jialin Wang, Wenbin Ge, Yang Fan, Kai Dang, Mengfei Du, Xuancheng Ren, Rui Men, Dayiheng Liu, Chang Zhou, Jingren Zhou, and Junyang Lin.
\newblock Qwen2-vl: Enhancing vision-language model's perception of the world at any resolution, 2024{\natexlab{b}}.

\bibitem[Wang et~al.(2024{\natexlab{c}})Wang, Sun, Zhang, Tang, Liu, and Wang]{wang2024diffusion}
Wenxuan Wang, Quan Sun, Fan Zhang, Yepeng Tang, Jing Liu, and Xinlong Wang.
\newblock Diffusion feedback helps clip see better.
\newblock \emph{arXiv preprint arXiv:2407.20171}, 2024{\natexlab{c}}.

\bibitem[Wang et~al.(2024{\natexlab{d}})Wang, Zhang, Luo, Sun, Cui, Wang, Zhang, Wang, Li, Yu, et~al.]{emu3}
Xinlong Wang, Xiaosong Zhang, Zhengxiong Luo, Quan Sun, Yufeng Cui, Jinsheng Wang, Fan Zhang, Yueze Wang, Zhen Li, Qiying Yu, et~al.
\newblock Emu3: Next-token prediction is all you need.
\newblock \emph{arXiv preprint arXiv:2409.18869}, 2024{\natexlab{d}}.

\bibitem[Wang et~al.(2022)Wang, Yu, Yu, Dai, Tsvetkov, and Cao]{wang2022simvlm}
Zirui Wang, Jiahui Yu, Adams~Wei Yu, Zihang Dai, Yulia Tsvetkov, and Yuan Cao.
\newblock Sim{VLM}: Simple visual language model pretraining with weak supervision.
\newblock In \emph{International Conference on Learning Representations}, 2022.

\bibitem[X.ai(2024)]{VLM:Grok-1.5}
X.ai.
\newblock Grok-1.5 vision preview.
\newblock \url{https://x.ai/blog/grok-1.5v}, 2024.

\bibitem[Xiao et~al.(2018)Xiao, Liu, Zhou, Jiang, and Sun]{xiao2018upernet}
Tete Xiao, Yingcheng Liu, Bolei Zhou, Yuning Jiang, and Jian Sun.
\newblock Unified perceptual parsing for scene understanding.
\newblock In \emph{Proceedings of the European conference on computer vision (ECCV)}, pages 418--434, 2018.

\bibitem[Yu et~al.(2022)Yu, Wang, Vasudevan, Yeung, Seyedhosseini, and Wu]{yu2022coca}
Jiahui Yu, Zirui Wang, Vijay Vasudevan, Legg Yeung, Mojtaba Seyedhosseini, and Yonghui Wu.
\newblock Coca: Contrastive captioners are image-text foundation models.
\newblock \emph{Transactions on Machine Learning Research}, 2022.

\bibitem[Yu et~al.(2024)Yu, Sun, Zhang, Cui, Zhang, Cao, Wang, and Liu]{yu2024capsfusion}
Qiying Yu, Quan Sun, Xiaosong Zhang, Yufeng Cui, Fan Zhang, Yue Cao, Xinlong Wang, and Jingjing Liu.
\newblock Capsfusion: Rethinking image-text data at scale.
\newblock In \emph{Proceedings of the IEEE/CVF Conference on Computer Vision and Pattern Recognition}, pages 14022--14032, 2024.

\bibitem[Yu et~al.(2023)Yu, Yang, Li, Wang, Lin, Liu, Wang, and Wang]{mmvet}
Weihao Yu, Zhengyuan Yang, Linjie Li, Jianfeng Wang, Kevin Lin, Zicheng Liu, Xinchao Wang, and Lijuan Wang.
\newblock Mm-vet: Evaluating large multimodal models for integrated capabilities.
\newblock \emph{arXiv preprint arXiv:2308.02490}, 2023.

\bibitem[Yuksekgonul et~al.(2022)Yuksekgonul, Bianchi, Kalluri, Jurafsky, and Zou]{yuksekgonul2022aro}
Mert Yuksekgonul, Federico Bianchi, Pratyusha Kalluri, Dan Jurafsky, and James Zou.
\newblock When and why vision-language models behave like bags-of-words, and what to do about it?
\newblock \emph{arXiv preprint arXiv:2210.01936}, 2022.

\bibitem[Zhai et~al.(2023)Zhai, Mustafa, Kolesnikov, and Beyer]{zhai2023siglip}
Xiaohua Zhai, Basil Mustafa, Alexander Kolesnikov, and Lucas Beyer.
\newblock Sigmoid loss for language image pre-training.
\newblock In \emph{Proceedings of the IEEE/CVF international conference on computer vision}, pages 11975--11986, 2023.

\bibitem[Zhang et~al.(2025)Zhang, Li, Huang, Li, Lei, Deng, Chen, Ji, and Feng]{zhangta2025pixelsail}
Tao Zhang, Xiangtai Li, Zilong Huang, Yanwei Li, Weixian Lei, Xueqing Deng, Shihao Chen, Shunping Ji, and Jiashi Feng.
\newblock Pixel-sail: Single transformer for pixel-grounded understanding.
\newblock \emph{arXiv}, 2025.

\bibitem[Zhang et~al.(2024)Zhang, Huang, Ma, Li, Luo, Xie, Qin, Luo, Li, Liu, et~al.]{RAM}
Youcai Zhang, Xinyu Huang, Jinyu Ma, Zhaoyang Li, Zhaochuan Luo, Yanchun Xie, Yuzhuo Qin, Tong Luo, Yaqian Li, Shilong Liu, et~al.
\newblock Recognize anything: A strong image tagging model.
\newblock In \emph{Proceedings of the IEEE/CVF Conference on Computer Vision and Pattern Recognition}, pages 1724--1732, 2024.

\bibitem[Zhou et~al.(2017)Zhou, Zhao, Puig, Fidler, Barriuso, and Torralba]{zhou2017ade20k}
Bolei Zhou, Hang Zhao, Xavier Puig, Sanja Fidler, Adela Barriuso, and Antonio Torralba.
\newblock Scene parsing through ade20k dataset.
\newblock In \emph{Proceedings of the IEEE conference on computer vision and pattern recognition}, pages 633--641, 2017.

\bibitem[Zhu et~al.(2023)Zhu, Chen, Shen, Li, and Elhoseiny]{zhu2023minigpt}
Deyao Zhu, Jun Chen, Xiaoqian Shen, Xiang Li, and Mohamed Elhoseiny.
\newblock Minigpt-4: Enhancing vision-language understanding with advanced large language models.
\newblock \emph{arXiv preprint arXiv:2304.10592}, 2023.

\end{thebibliography}
}

\end{document}